%% file: main.tex
\documentclass[10pt,twocolumn,letterpaper]{article}

\usepackage{wacv}              

\input{paper/preamble}

\definecolor{wacvblue}{rgb}{0.21,0.49,0.74}

\definecolor{HeaderGray}{gray}{1.00}
\definecolor{ResultGray}{gray}{0.86}
\usepackage[pagebackref,breaklinks,colorlinks,allcolors=wacvblue]{hyperref}

\def\wacvPaperID{1948} 
\def\confName{WACV}
\def\confYear{2027}

\title{Learning Lane-Structured Conditioning Fields for Online HD Mapping}

\author{
Andrea Bertogalli$^{1, \dagger}$ \quad
Anthony Dell'Eva$^{3}$ \quad
Marcello Ceresini$^{1}$ \quad
Federico Pirazzoli$^{1}$ \\
Lorenzo Cipelli$^{1}$ \quad
Filippo D'Addeo$^{2}$ \quad
Nicola Marinello$^{3}$ \quad
Massimo Bertozzi$^{1}$\\[0.5em]
$^{1}$University of Parma, Italy \quad
$^{2}$University of Bologna, Italy \quad
$^{3}$VisLab, Italy\\[0.3em]
}

\begin{document}
\maketitle

\begingroup
\renewcommand{\thefootnote}{\textbf{\textdagger}}
\footnotetext{Corresponding author: \texttt{andrea.bertogalli@unipr.it}}
\endgroup

\input{paper/sec/0_abstract}
\input{paper/sec/1_intro}
\input{paper/sec/2_related}
\input{paper/sec/3_method}
\input{paper/sec/4_experiments}
\input{paper/sec/5_conclusion}


{
    \small
    \bibliographystyle{ieeenat_fullname}
    \bibliography{paper/main}
}

\end{document}


\maketitle

\begingroup
\renewcommand{\thefootnote}{\textbf{\textdagger}}
\footnotetext{Corresponding author: \texttt{andrea.bertogalli@unipr.it}}
\endgroup

\appendix

\input{supplementary/sec/0_overview}
\input{supplementary/sec/1_derivation}
\input{supplementary/sec/2_sigma_ablation}

\input{supplementary/sec/2.5_implementation_details}
\input{supplementary/sec/3_loss_functions}
\input{supplementary/sec/4_variance}
\input{supplementary/sec/5_attention}
\input{supplementary/sec/6_modulation}
\input{supplementary/sec/7_qualitative}
\input{supplementary/sec/8_inference}
\input{supplementary/sec/9_gradients}
\input{supplementary/sec/big_images}

\clearpage
{
    \small
    \bibliographystyle{ieeenat_fullname}
    \bibliography{supplementary/main}
}

%% file: paper/preamble.tex
\usepackage{multirow}
\usepackage{colortbl}
\usepackage{float}
\usepackage{makecell}

\usepackage{multirow}
\usepackage[table]{xcolor}
\usepackage{hhline}
\usepackage{tabularx}

\usepackage[table]{xcolor}
\usepackage{adjustbox}

\definecolor{TodoBlue}{RGB}{30,120,200}
\definecolor{TodoBlueBg}{RGB}{225,240,255}

%% file: paper/sec/0_abstract.tex
\begin{abstract}
High-definition maps provide critical structural information for downstream autonomous driving tasks.
Recent methods increasingly rely on external priors to improve prediction accuracy and robustness.
Yet, existing prior representations expose a trade-off: dense rasters are easy to integrate but ignore certain map geometric properties, while sparse vectors preserve geometric structure but require dedicated encoders and are difficult to plug into existing architectures.
Moreover, external priors can be inaccurate, costly to acquire, or difficult to maintain, motivating a structural representation that can provide geometric guidance while remaining learnable and broadly applicable.
We address this limitation with the Lane Field, a dense BEV-aligned geometric representation that encodes local lane structure at each spatial location and can be either predicted from onboard observations or constructed from external map priors.
Building on this representation, we introduce a plug-and-play framework that enhances online HD mapping through two complementary mechanisms: BEV feature modulation, which injects lane-structured cues into latent BEV features, and deformable attention guidance, which steers query sampling toward the underlying lane geometry.
We primarily evaluate our method on \mbox{OpenLane-V2}, showing that conditioning BEV features and guiding deformable attention with local lane geometry enables our framework to consistently improve existing methods and achieve state-of-the-art performance on lane segment perception. Project page: \href{https://andberto.github.io/LaneField/}{https://andberto.github.io/LaneField/} 
\end{abstract}

%% file: paper/sec/1_intro.tex
\begin{figure}[t]
    \centering
    \includegraphics[width=\columnwidth]{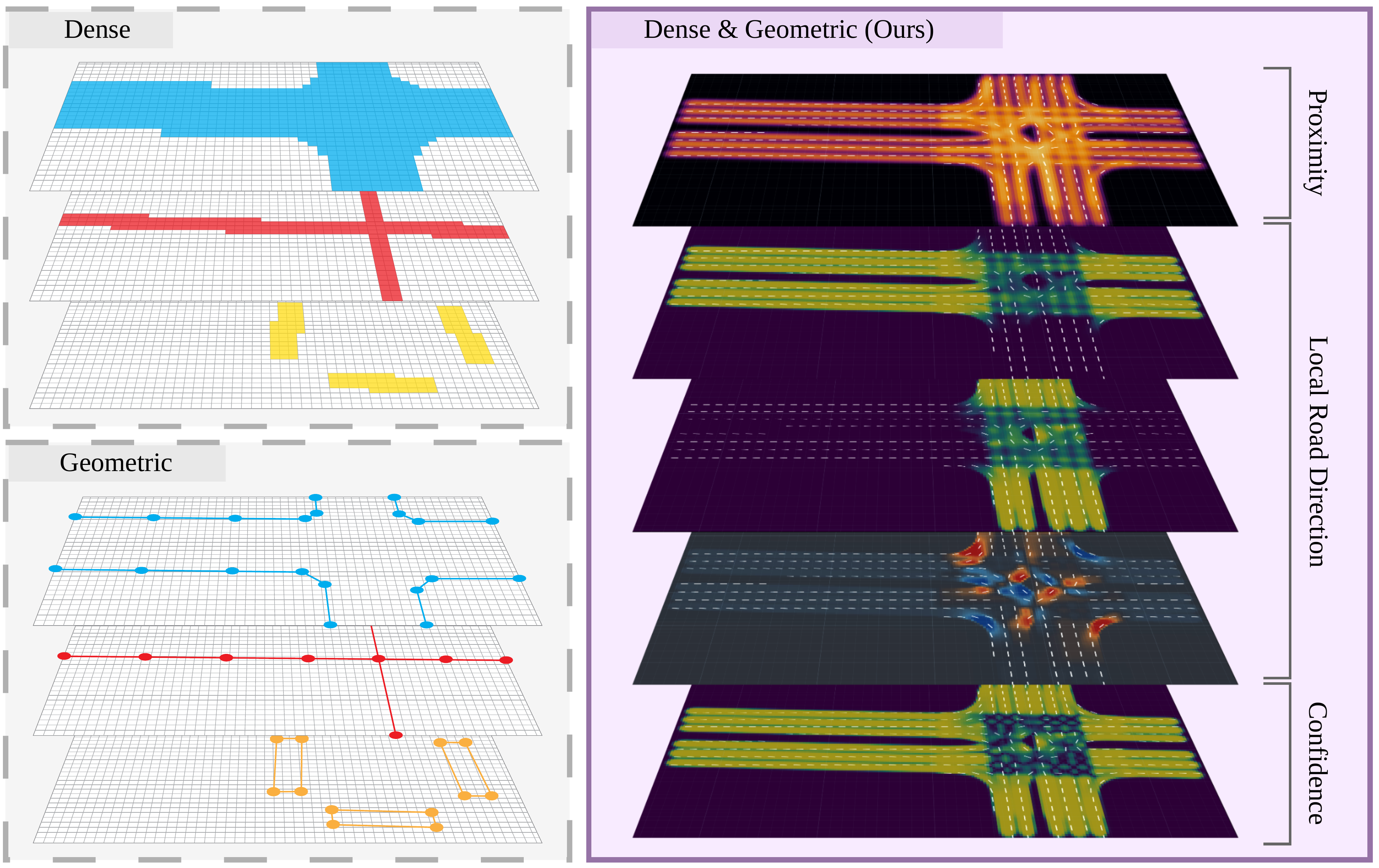}
\caption{
Comparison between classical map representations commonly used as additional supervision or external priors and our proposed representation, the Lane Field. 
Our method elegantly combines the geometric meaning of vectorized representations with a dense BEV-aligned spatial structure.}
\label{fig:introduction}
\end{figure}

\section{Introduction}
\label{sec:intro}
Understanding road structure is a fundamental requirement for autonomous driving. 
High-definition (HD) maps capture this structure through precise vectorized representations of key map elements, such as centerlines, lane dividers, road boundaries, and pedestrian crossings~\cite{liao2025maptrv2, wilson2023argoverse, caesar2020nuscenes}.
Beyond geometry alone, modern HD map formulations also encode relational information, including lane connectivity and associations between traffic lights, traffic signs, and the lanes they regulate~\cite{wang2023openlane, li2023lanesegnet}.
This has motivated recent methods to move beyond isolated map elements toward lane segment perception and topology reasoning, providing a more complete and structured road scene representation. These geometric and relational cues provide essential context for downstream tasks such as motion planning.

Recent advances in deep learning, onboard sensing and hardware acceleration have shifted HD map construction from an offline process toward online HD mapping~\cite{jsan14010015}, where map elements are inferred directly from onboard observations in real time. Early approaches formulated online HD mapping as a BEV semantic segmentation problem~\cite{li2021hdmapnet, philion2020lift}. 
However, rasterized representations do not explicitly model instances and tend to discard the structured geometry of map elements, which carries rich information about road layout and spatial relationships. 
Recent vectorized HD mapping methods address this limitation by decoding map elements directly as geometric primitives, such as polylines, often through DETR-style query-based architectures~\cite{carion2020end, liao2022maptr, liao2025maptrv2, liu2023vectormapnet}. 
While this paradigm better preserves geometric coherence, its query-based decoding process involves only a limited subset of the dense BEV feature grid when predicting map elements.
As a result, the BEV representation is supervised only through sparse query interactions, which can weaken its spatial structure and make it harder to capture fine-grained geometric cues. 

Therefore, we argue that an intermediate geometric support representation can help bridge this gap by preserving lane structure while providing dense, spatially aligned guidance over the BEV feature grid.
To this end, we introduce the \textit{Lane Field} (LF), a dense geometric representation defined on the BEV grid, shown in \Cref{fig:introduction}. 
It encodes local unoriented lane direction at each spatial location, converting sparse vector map geometry into a spatially dense signal that can support BEV-based decoding. 
We demonstrate its versatility by formulating LF prediction as an auxiliary supervised task, using ground truth data generated analytically from vectorized maps. 
We further integrate the LF into existing architectures through two plug-and-play enhancements: a BEV feature modulation block and a deformable attention guidance mechanism. 
In contrast to a simple rasterized approach, the LF makes these two integrations possible within a single representation, since it provides both dense BEV-aligned cues for feature conditioning and explicit geometric information for attention guidance.

\noindent In summary, the main contributions of this paper are:

\begin{itemize}
    \item We introduce the Lane Field, a dense BEV-aligned representation that encodes lane geometry at each spatial location. We provide a novel analytic geometric construction to derive ground truth fields from vector map annotations, enabling direct supervision within modern mapping pipelines.

    \item We propose a plug-and-play BEV feature conditioning module that uses the Lane Field to inject lane-structured geometric cues into latent BEV features for downstream map decoding.

    \item We introduce a plug-and-play deformable attention guidance mechanism that exploits the Lane Field to condition the sampling locations, steering the decoder toward meaningful regions.
\end{itemize}

%% file: paper/sec/2_related.tex
\section{Related Works}
\label{sec:related}
\subsection{Bird’s-Eye-View Perception}
As HD map construction is typically formulated in \mbox{Bird's-Eye} View (BEV), most online HD mapping architectures rely on a perspective-view (PV)-to-BEV encoder to transform image features into a unified BEV representation. Philion and Fidler~\cite{philion2020lift} propose one of the first camera-only approaches to construct dense BEV representations, by lifting image features into 3D frustums and splatting them onto a rasterized BEV grid. BEVFormer~\cite{li2024bevformer} introduces a transformer-based BEV encoder that aggregates multi-camera and temporal information into a unified BEV feature representation for downstream tasks. Following this idea, other works enhancing BEV feature construction were proposed. BEVFormerV2~\cite{yang2023bevformer} extends BEVFormer with perspective view supervision, providing additional training guidance before BEV feature aggregation. BEVFusion~\cite{liu2023bevfusion} explores the integration of LiDAR information into BEV perception, while SimpleBEV~\cite{harley2023simple} adopts an inverse perspective mapping feature sampling strategy, projecting 2D image features onto predefined BEV anchor point.

\subsection{HD Map Construction}
\noindent\textbf{Map Element Detection.}
Early online HD mapping approaches represented map elements as rasterized BEV masks, reducing map construction to dense semantic segmentation followed by vectorization, as in HDMapNet~\cite{li2021hdmapnet}. Later methods directly decoded map elements as geometric primitives: VectorMapNet~\cite{liu2023vectormapnet} predicts vectorized polylines, while MapTR and MapTRv2~\cite{liao2022maptr,liao2025maptrv2} established DETR-style query-based decoding. Alternative representations, such as PivotNet~\cite{ding2023pivotnet}, model map elements through pivot-based point sets. Mask2Map~\cite{choi2024mask2map} and MGMap~\cite{liu2024mgmap} leverage learned BEV masks to inject semantic and geometric cues into vector decoding, whereas MapQR~\cite{liu2024leveraging} enhances point-set queries through scatter-and-gather interactions. Subsequent methods model geometric relationships among map elements~\cite{zhang2024online} and leverage temporal information across frames~\cite{yuan2024streammapnet,wang2024stream}.
\\\\\noindent\textbf{Centerline Perception.}
Recent benchmarks and methods extend online HD mapping beyond isolated map element prediction toward lane-level map construction. \mbox{OpenLane-V2}~\cite{wang2023openlane} requires models to recover lane-level geometry together with relational information. Several methods combine centerline perception with lane-lane topology reasoning: TopoNet~\cite{li2026graph} uses graph neural networks to model lane-lane and lane-traffic relations, TopoPoint~\cite{fu2026topopoint} focuses on endpoint localization for connectivity, TopoMLP~\cite{wu2024topomlp} injects coordinate information into relational reasoning, and TopoFormer \cite{lv2025t2sg} introduces a unified graph representation to explicitly model lanes.
\\\\\noindent\textbf{Lane Segment Perception.} LaneSegNet~\cite{li2023lanesegnet} introduces the lane segment perception task, where each lane segment jointly encodes a centerline, its associated boundaries and their semantic attributes, and directed connectivity, providing a more complete representation of road scene for both map perception and topology reasoning.
Building on this representation, subsequent works extend lane-segment perception through complementary strategies. TopoLogic~\cite{fu2024topologic} combines geometric and semantic lane cues for topology reasoning, while Topo2Seq~\cite{yang2025topo2seq} formulates topology prediction as sequence decoding.

A common architectural component shared by most methods is the use of the underlying Deformable DETR mechanism \cite{zhu2020deformable}. While this mechanism improves convergence and reduces computational cost, it also limits the amount of dense spatial supervision propagated through the BEV feature grid. As a result, the learned latent representation may exhibit weaker spatial structure and overlook fine-grained geometric details that are important for accurate map construction. Our framework addresses these limitations through two complementary mechanisms built upon a single dense geometric representation, termed Lane Field. First, it reinforces the spatial structure of latent BEV features by leveraging the LF for feature modulation. Second, it guides deformable attention toward geometrically meaningful regions, aligning query sampling with the underlying lane structure.

\subsection{Priors and Additional Supervision Methods}
Incorporating external priors can provide additional structural cues and improve online HD mapping, particularly under severe occlusions. P-MapNet~\cite{jiang2024p} uses rasterized SD maps as external priors to improve HD map construction and enable longer-range perception. SMERF~\cite{luo2024augmenting} represents SD maps as polyline sequences and encodes them with a Transformer encoder. Gao \etal~\cite{gao2024complementing} complement onboard sensor features with satellite maps, while Jia \etal~\cite{jia2025enhancing} exploit crowdsourced vehicle trajectories, encoding them as rasterized heatmaps and vectorized tokens.

Directional cues and vector fields have also been explored in other maps related tasks~\cite{batra2019improved, li2022reconstruct, liang2021road, liang2019convolutional}. However, our analytic Lane Field construction, combined with the proposed conditioning mechanisms and its flexibility across learned and prior-based settings, has not been previously explored in online HD mapping.

AugMapNet~\cite{monninger2026augmapnet} introduces an intermediate dense augmentation step by injecting features from a raster map prediction into the latent BEV grid. Building on the idea of intermediate supervision, our Lane Field analytically densifies vectorized lane geometry into a BEV-aligned representation encoding direction, proximity, and directional confidence, and exploits it for both BEV feature conditioning and deformable attention guidance. Importantly, the LF can be either learned from onboard observations or constructed from any polyline-based prior (\eg SD maps) through the same analytic procedure.

%% file: paper/sec/3_method.tex
\section{Method}
\begin{figure*}[t]
    \centering
    \includegraphics[width=\textwidth]{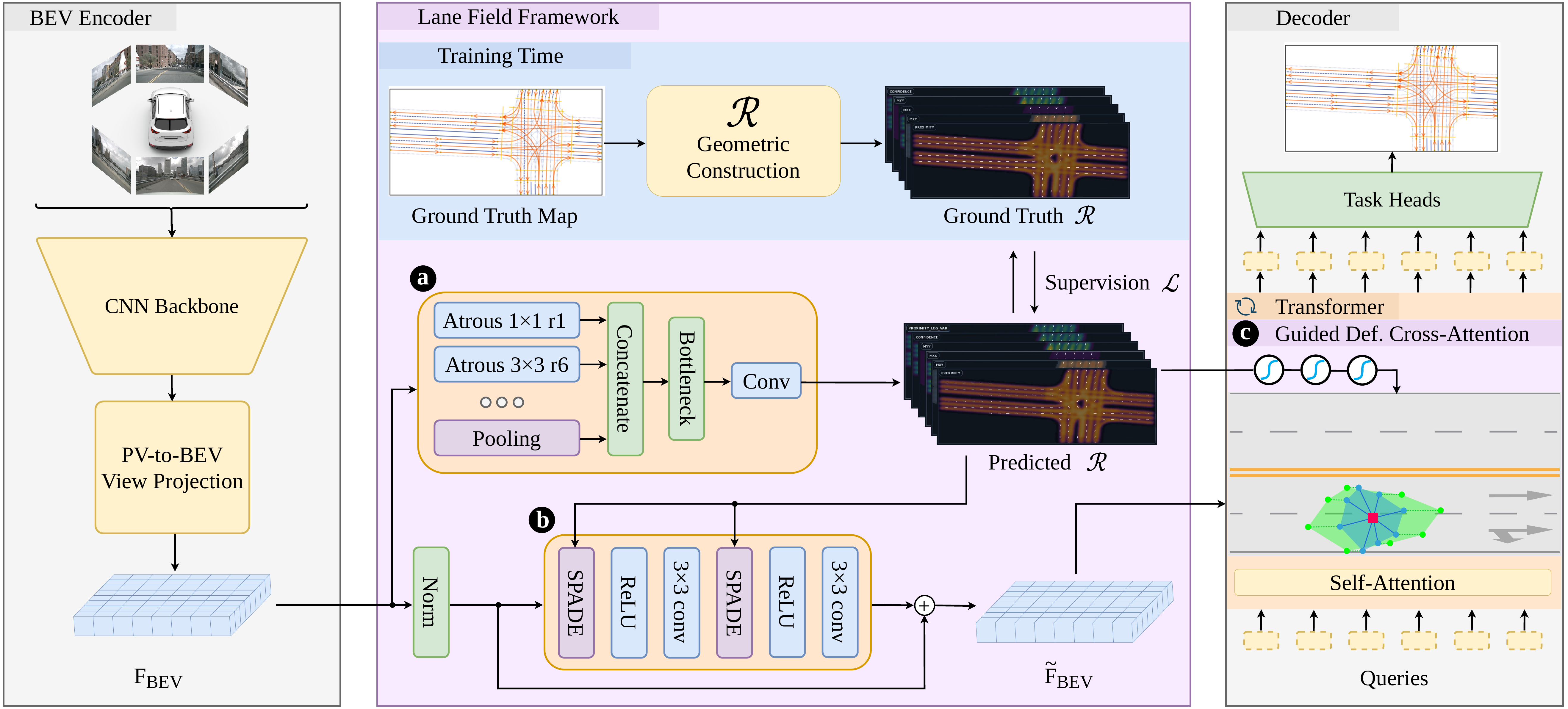}
\caption{
Overview of the proposed framework. 
Starting from \(F_{\mathrm{BEV}}\), a prediction head estimates the Lane Field \textbf{(a)}, supervised by ground-truth fields geometrically constructed from vectorized map annotations. 
The predicted LF is then used to modulate the BEV features through a SPADE-like residual block \textbf{(b)} and to guide deformable attention sampling \textbf{(c)} toward geometrically meaningful roadway regions.
}
    \label{fig:architecture}
\end{figure*}
\subsection{Framework overview}
Given a multi-camera setup that provides multi-view images of the scene, our framework aims to construct the corresponding HD map in Bird’s-Eye View. Although our implementation uses camera inputs, the proposed framework is agnostic to the sensing modality and BEV encoder, as it operates on latent BEV features independently of how they are extracted from raw sensor measurements. To estimate the HD map, the pipeline follows a standard encoder-decoder design. A shared image backbone extracts multi-view features from the input images, while a view transformation module aggregates them and projects them into a BEV feature grid of predefined spatial size ${H\times W}$. We denote the feature derived after the encoder as ${F_{BEV}\in \mathbb{R}^{H\times W\times C}}$ where $C$ is the number of channels. The BEV feature grid is then processed by a DETR-like decoder~\cite{zhu2020deformable}, where learned instance queries interact with BEV features to predict vectorized map element instances. Our framework enhances this standard pipeline through two complementary mechanisms. First, the proposed LF is used to modulate the BEV feature grid $F_{BEV}$, conditioning the latent representation on dense lane-geometry information. Second, the same representation guides deformable attention sampling, helping the decoder extract information from geometrically relevant regions in the BEV space. An overview of the framework is presented in \Cref{fig:architecture}.

\subsection{The Lane Field}
\label{sec:road_field}
While previous methods exploit dense BEV supervision or directional representations, we focus on analytically densifying vectorized lane geometry into a directional tensor field encoding local direction, proximity, and confidence. The resulting Lane Field serves as an intermediate geometric representation for both BEV feature conditioning and explicit deformable attention guidance. The LF, denoted as \(\mathbf{\mathcal{R}}  \in \mathbb{R}^{H \times W \times 5}\), shares the same spatial resolution as \(F_{\mathrm{BEV}}\) and encodes local geometric lane information at each BEV location through five channels. Following \cite{li2023lanesegnet}, each lane is described by a centerline polyline \(C=\{C_k\}_{k=1}^{n}\), where \(C_k= (x_k,y_k)\in\mathbb{R}^2\) and \(n\) is the number of points of the polyline. At each centerline point, left and right offsets define the lane boundaries, from which we estimate the local lane width \(w_k\).

Each centerline polyline is decomposed into \(n-1\) consecutive segments.
For segment \(i\), with endpoints \(A_i=C_k\) and \(B_i=C_{k+1}\), we define its unit tangent \(\hat{\mathbf{t}}_i\) and corresponding lane width \(\bar{w}_i\) as
\begin{equation}
\hat{\mathbf{t}}_i
=
\frac{\overrightarrow{A_iB_i}}{\ell_i}
=
\frac{B_i-A_i}{\ell_i},
\qquad
\bar{w}_i
=
\frac{w_k+w_{k+1}}{2},
\end{equation}
where \(\ell_i=\|\overrightarrow{A_iB_i}\|\) is the segment length.  Given the unit tangent \(\hat{\mathbf{t}}_i\), we encode the unoriented direction of segment \(i\) through the outer product \(\hat{\mathbf{t}}_i \otimes \hat{\mathbf{t}}_i = \hat{\mathbf{t}}_i \hat{\mathbf{t}}_i^\top \in \mathbb{R}^{2\times 2}\). To distribute segment-level geometry over the dense BEV grid, each segment of each lane centerline contributes a smooth weight to every spatial location. For segment \(i\) and BEV location \(p\), we define this contribution as the line integral of a Gaussian kernel along the segment:
\begin{equation}
I_i(p)
=
\ell_i
\int_0^1
\exp\left(
-\frac{\|p - A_i - t(B_i-A_i)\|^2}{2\sigma_i^2}
\right) dt,
\label{eq:integral}
\end{equation}
where \(t\in[0,1]\) parametrizes the segment, with \(t=0\) corresponding to \(A_i\) and \(t=1\) to \(B_i\). 

\noindent This integral has a closed-form solution:
\begin{equation}
I_i(p)
=
\frac{\sigma_i\sqrt{2\pi}}{2}
\exp\left(
-\frac{d_{\perp,i}^2}{2\sigma_i^2}
\right)
\left[
\operatorname{erf}(u_{1,i}) - \operatorname{erf}(u_{0,i})
\right],
\end{equation}
where \(d_{\perp,i}\) is the perpendicular distance from \(p\) to the line supporting the segment, and \(u_0,u_1\) are the transformed bounds associated
with the finite segment endpoints \footnote{\(u_{0,i}=-\frac{(p-A_i)^\top\overrightarrow{A_iB_i}}{\sqrt{2}\sigma_i\ell_i}\), 
\(u_{1,i}=u_{0,i}+\frac{\ell_i}{\sqrt{2}\sigma_i}\).} .
We leave the full derivation in the supplementary material.

The spread parameter \(\sigma_i\) controls the spatial extent of each segment contribution over the BEV grid. 
It can be set globally or adapted to the local lane width as
\begin{equation}
\sigma_i = f \cdot \bar{w}_i .
\end{equation}
The factor \(f\) determines the lateral spread of the Gaussian kernel, yielding a value of \(\exp(-1/(8f^2))\) at the lane boundary, \ie at distance \(\bar{w}_i/2\) from the centerline.

We can now define the Lane Field \(\mathbf{\mathcal{R}}\). 
The first three channels, \(M_{xx}\), \(M_{xy}\), and \(M_{yy}\), encode the local unoriented lane direction. 
To obtain them, we define at each BEV location \(p\) a \(2\times2\) symmetric matrix by accumulating the outer product direction of all segments, weighted by their contribution \(I_i(p)\):
\begin{equation}
\mathbf{M}(p)
=
\sum_i I_i(p)\, \hat{\mathbf{t}}_i \otimes \hat{\mathbf{t}}_i
=
\begin{pmatrix}
M_{xx}(p) & M_{xy}(p) \\
M_{xy}(p) & M_{yy}(p)
\end{pmatrix}.
\end{equation}
 As additional channel, we define a proximity measure \(\mathbf{P}(p)\), which indicates how strongly each BEV location is supported by nearby lanes. 
It is obtained by accumulating the segment contributions:
\begin{equation}
    \mathbf{P}(p) = \sum_i I_i(p).
\end{equation}

Finally, since the local direction may be ambiguous in regions such as intersections, we introduce a direction confidence channel \(\mathbf{C}(p)\), which measures the reliability of the directional tensor at each BEV location.
The confidence is derived from the anisotropy of \(\mathbf{M}(p)\). When a single dominant direction is present, the tensor is highly anisotropic, indicating a reliable orientation. Conversely, in ambiguous regions where multiple directions contribute similarly, the anisotropy decreases, reflecting lower reliability. Let the eigenvalues of \(\mathbf{M}(p)\) be
\[
\lambda_{1,2}(p)
=
\frac{\operatorname{tr}(p) \pm \Delta(p)}{2},
\]
where \(\operatorname{tr}(p)=M_{xx}(p)+M_{yy}(p)\) and
\[
\Delta(p)=
\sqrt{
\left(M_{xx}(p)-M_{yy}(p)\right)^2
+
4M_{xy}(p)^2
}.
\]
We define the direction confidence channel as
\begin{equation}
\mathbf{C}(p)
=
\frac{\lambda_1(p)-\lambda_2(p)}
{\lambda_1(p)+\lambda_2(p)}
=
\frac{\Delta(p)}{\operatorname{tr}(p)}.
\end{equation}
The LF at BEV location \(p\) is therefore defined as
\begin{equation}
\mathbf{\mathcal{R}}(p)=
\left[
\mathbf{M}_{xx}(p),\,
\mathbf{M}_{xy}(p),\,
\mathbf{M}_{yy}(p),\,
\mathbf{P}(p),\,
\mathbf{C}(p)
\right].
\end{equation}
The directional channels and the confidence are soft-masked by the proximity channel to suppress regions with insufficient road evidence. 
After normalization, \(\mathbf{M}_{xx}\), \(\mathbf{M}_{yy}\), \(\mathbf{P}\), and \(\mathbf{C}\) lie in \([0,1]\), while \(\mathbf{M}_{xy}\in[-\frac{1}{2},\frac{1}{2}]\).
For clarity, we omit minor implementation details related to numerical stability, which are provided in the supplementary material.

\subsection{Learning the Lane Field}
Although \(\mathcal{R}\) can be analytically constructed from polyline-based priors such as SD maps, these can be hard to retrieve and integrate.
We therefore formulate LF estimation as an auxiliary dense prediction task, using ground truth data constructed from HD map annotations as supervision. To this end, we adopt a DeepLabV3-like head~\cite{chen2017rethinking}, which takes \(F_{\mathrm{BEV}}\) as input and produces a dense prediction of \(\mathcal{R}\). 
\\\\
\noindent\textbf{Losses.}
We supervise LF prediction with multiple losses, each targeting a different property of the representation. First, we apply a Smooth-\(L_1\) regression  loss \(\mathcal{L}_{R}\) \cite{girshick2015fast} to the directional tensor channels \(\mathbf{M}_{xx}\), \(\mathbf{M}_{xy}\), and \(\mathbf{M}_{yy}\). This term encourages the network to accurately predict the local geometric structure of the lane at each BEV location. 

Second, inspired by \cite{mathieu2015deep}, we apply an \(L_1\) loss \(\mathcal{L}_{G}\) to the spatial gradients of the directional channels, computed through horizontal and vertical finite differences, encouraging sharper edges. 

Finally, following \cite{gu2024producing}, we supervise the proximity channel with a Laplacian \(\beta\)-NLL loss \(\mathcal{L}_{L}\) by predicting both its mean and log-scale. 
This formulation provides an uncertainty estimate in addition to the proximity prediction, which can be used as a reliability cue by the BEV feature modulation block and the deformable attention guidance mechanism. Since the proximity channel can be zero over most of the BEV grid, the Laplacian loss can be dominated by background locations. We therefore introduce a weighting term that emphasizes high-proximity regions, encouraging the model to better learn road-supported areas. The resulting per-location \(\beta\)-NLL loss is defined as:
\begin{equation}
\ell^{L}_{i}
=
\left(
\log(2 b_i)
+
\frac{\bigl|\hat{\mu}_i-\mu_i\bigr|}{b_i}
\right)
\cdot
\lfloor b_i \rfloor^{\beta}
\cdot
w^{\mathrm{prox}}_{i},
\end{equation}
where $b_i$ is the scale, $\hat{\mu}_i$ is the predicted proximity mean, $\mu_i$ is the ground truth proximity, $\lfloor \cdot \rfloor$ denotes the stop gradient operation, and
\begin{equation}
w^{\mathrm{prox}}_{i}
=
w_{\min}
+
(1-w_{\min})\,\sigma\!\left(\alpha\cdot(\mu_i-\tau)\right)
\end{equation}
with $\sigma$ the sigmoid function, $w_{min}$, $\alpha$ and $\tau$ are defined in the supplementary material. Note that we do not supervise $\mathbf{C}(p)$ since we derive it analytically from predicted channels. The total loss is a weighted sum of the three losses presented above:
\begin{equation}
    \mathcal{L}
    =
    w_R \cdot \mathcal{L}_{R}
    +
    w_G \cdot \mathcal{L}_{G}
    +
    w_L \cdot \mathcal{L}_{L}.
\end{equation}

\subsection{BEV Feature Modulation}
\label{sec:modulation}
\begin{figure}[t]
    \centering
    \includegraphics[width=\columnwidth]{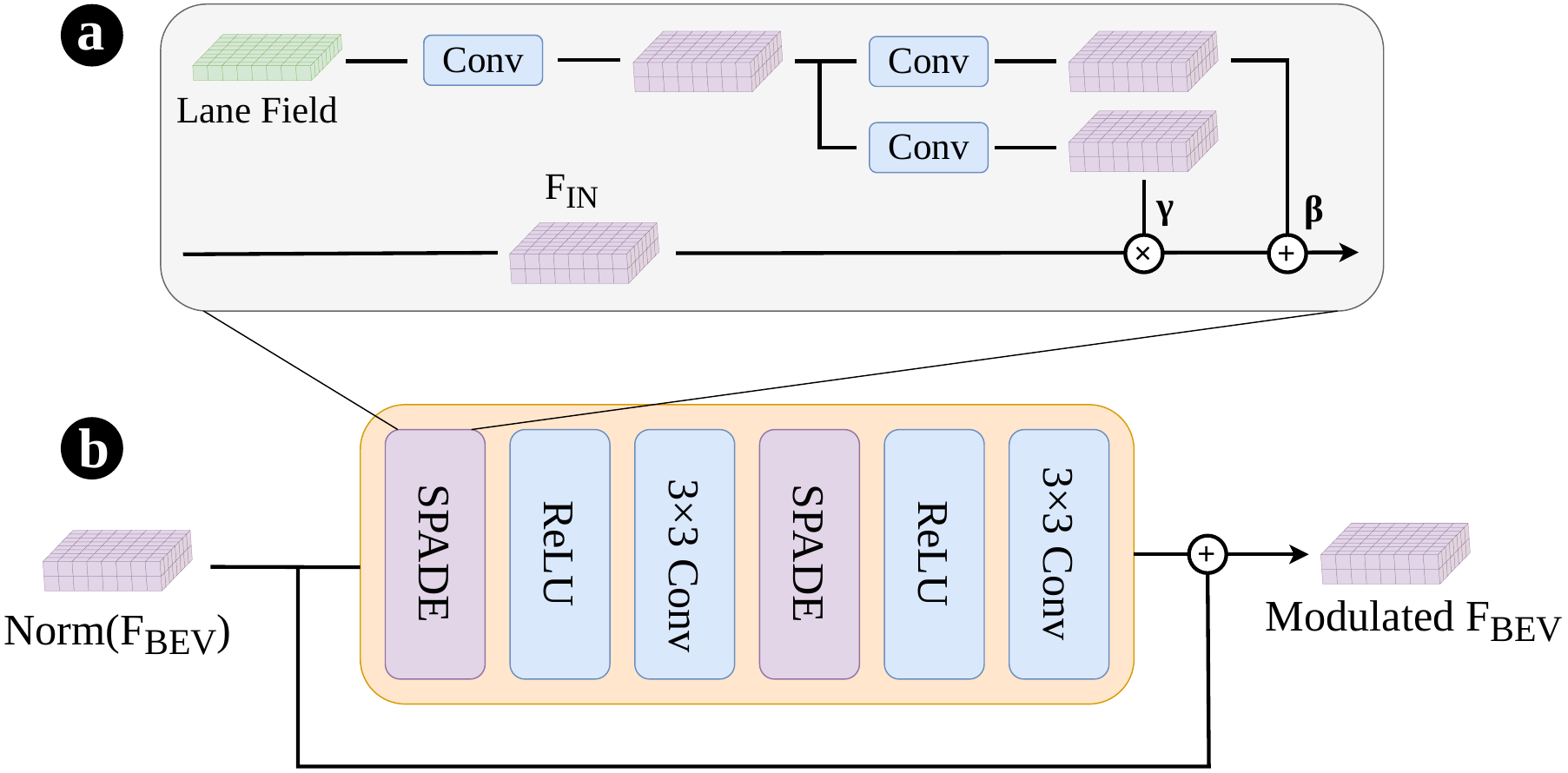}
\caption{
\textbf{(a)} SPADE-like modulation block. Given the predicted Lane Field, we project it into a shared latent representation, which is then processed by two separate convolutional branches to predict the modulation parameters \(\gamma\) and \(\beta\).
\textbf{(b)} Residual BEV modulation block. The modulation operation is embedded into a residual block, which can be stacked to progressively refine the BEV features.
}
        \label{fig:SPADE}
\end{figure}
We introduce a BEV feature modulation block to inject LF information into the latent BEV representation.
Inspired by SPADE \cite{park2019semantic}, which uses semantic maps to generate spatially-varying affine parameters for feature modulation, we condition the BEV features on the predicted LF.
This allows the network to adapt the latent representation according to the local lane geometry encoded by the field. As shown in Figure \ref{fig:SPADE}a, we adapt the SPADE mechanism to our method. 
Starting from the predicted LF, we first compute a shared latent representation, which is then processed by two separate convolutional branches to predict spatially varying affine parameters 
\(\gamma,\, \beta \in \mathbb{R}^{H \times W \times C}\), matching the shape of \(F_{\mathrm{BEV}}\). The output features are computed as
\(
F_{\mathrm{OUT}}
=
\gamma \odot F_{\mathrm{IN}}
+
\beta,
\)
where \(\odot\) denotes element-wise multiplication.
We implement this operation within a residual BEV refinement block, illustrated in \Cref{fig:SPADE}b.
The block takes \(F_{\mathrm{BEV}}\) as input and outputs its spatially modulated counterpart  \(\Tilde{F}_{\mathrm{BEV}}\).

\subsection{Deformable Attention Guidance}
\label{sec:attention}
\begin{figure}[t]
    \centering
    \includegraphics[width=\columnwidth]{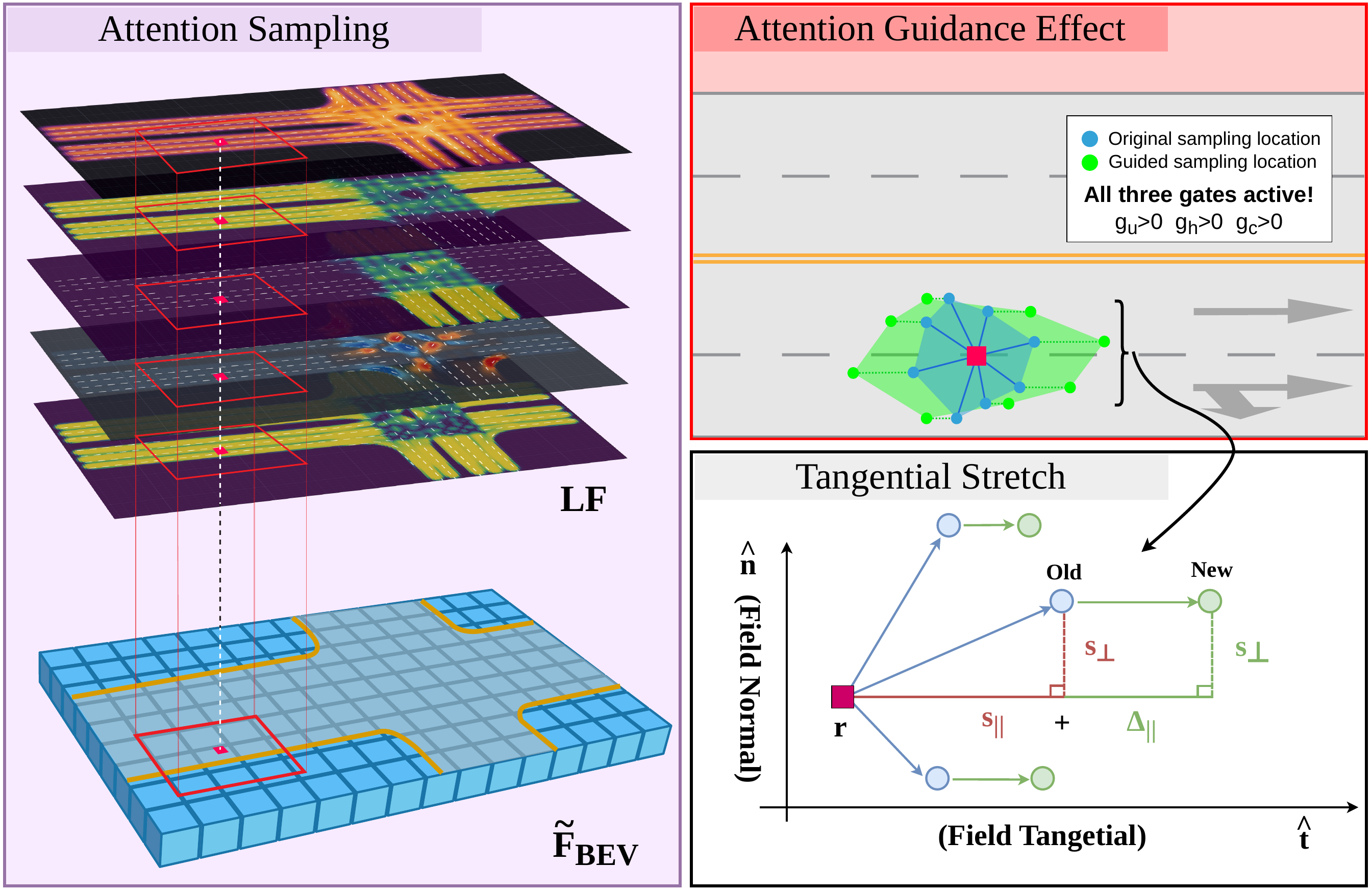}
\caption{
Effect of the Lane Field guidance mechanism on deformable attention sampling locations around a reference point. 
When all three gates are active, the sampling locations are steered along the local lane direction, yielding a lane aligned sampling pattern. The tangential stretch is defined in \Cref{eq:attn}.
}
        \label{fig:attention_guiding}
\end{figure}
Beyond BEV feature modulation, we exploit the LF to guide deformable cross-attention in query-based decoders. 
Existing HD mapping architectures commonly rely on deformable attention variants to capture the elongated structure of map elements~\cite{li2023lanesegnet,yuan2024streammapnet}. 
However, their sampling offsets are learned without an explicit conditioning signal describing the local lane geometry. 
We instead use the LF to make the sampling process lane aware by adapting the deformable attention pattern according to the local lane direction, as illustrated in \Cref{fig:attention_guiding}.

Given a reference point \(\mathbf r\), we bilinearly sample the LF at
\(\mathbf r\), obtaining the local \(2\times2\) directional tensor
\(\mathbf M(\mathbf r)\), and derive its unoriented dominant tangent
axis as the principal eigenspace of \(\mathbf M(\mathbf r)\). We choose
an arbitrary unit representative \(\hat{\mathbf t}\) of this axis and
its corresponding normal \(\hat{\mathbf n}\). Each learned deformable
attention sampling location \(\mathbf s\) is interpreted in this
lane-aligned coordinate frame, with signed tangential coordinate
\(s_{\parallel}\) and normal coordinate \(s_{\perp}\). Our guidance
preserves the normal component, while stretching the sampling location
along the local lane axis:
\begin{equation}
\mathbf{s}_{\mathrm{new}}
=
\mathbf{r}
+
\left[
1+
g_C(\mathbf r)g_U(\mathbf r)g_H(h)
(\Delta_{\parallel}-1)
\right]
s_{\parallel}\hat{\mathbf t}
+
s_{\perp}\hat{\mathbf n}.
\label{eq:attn}
\end{equation}
Here, \(g_C(\mathbf{r})\), \(g_U(\mathbf{r})\), and \(g_H(h)\) denote the
direction confidence gate, the uncertainty gate, and the learnable
per-head gate, respectively, while \(\Delta_\parallel\) is the maximum
tangential stretch factor. Since \(s_{\parallel}\) is signed in the same
local frame, samples on opposite sides of \(\mathbf r\) are pushed in
opposite tangential directions, making the formulation invariant to the
arbitrary sign of \(\hat{\mathbf t}\).
The direction confidence gate is derived from the LF confidence channel \(\mathbf{C}(\mathbf{r})\), and activates the guidance only when a dominant local direction can be reliably inferred. 
This avoids enforcing a wrong direction in ambiguous regions, such as intersections, where multiple lane directions coexist. 
The uncertainty gate suppresses the guidance in regions with high predicted uncertainty, while the learnable per-head gate allows each attention head to adaptively control the strength of the geometric guidance. More details can be found in the supplementary material.

%% file: paper/sec/4_experiments.tex
\section{Experiments}
\subsection{Datasets and Metrics}

\noindent\textbf{OpenLane-V2.} 
OpenLane-V2~\cite{wang2023openlane} is a benchmark for unified 3D HD map perception and topology reasoning, providing annotations for both lane segments and centerlines. Following the LaneSegNet~\cite{li2023lanesegnet} setting, we use subset A, built on Argoverse 2, which contains 1k scenes from six cities, with approximately 27k training frames and 4.8k validation frames. We evaluate lane-segment and centerline perception, together with lane-lane topology in their respective settings. Each lane segment is annotated as a directed 3D instance comprising a centerline and corresponding left and right boundaries, all represented as ordered point sequences. Boundaries additionally carry line-type labels (solid, dashed, or non-visible), while directed lane-lane adjacency annotations specify their topological relationships. For centerline perception, we use the provided centerline annotations and centerline-to-centerline relations.
\\\noindent\textbf{Evaluation metrics.}
We evaluate using the official OpenLane-V2 evaluation protocol. For lane segment perception, we report mAP, AP$_{ls}$, AP$_{ped}$, and TOP$_{lsls}$ at distance thresholds ${1.0, 2.0, 3.0}\mathrm{m}$, where TOP$_{lsls}$ measures topology prediction between lane segments. For centerline perception, we report DET$_l$ for centerline detection and TOP$_{ll}$ for centerline topology prediction with thresholds ${1.0, 2.0, 3.0}\mathrm{m}$. We also report the OLS measure only between DET$_l$ and TOP$_{ll}$ as in ~\cite{li2023lanesegnet}.

\subsection{Implementation details}
Unless otherwise specified, all models are trained on 8 NVIDIA RTX 4000 Ada Generation GPUs. To assess the relative impact of our method, we preserve the original configurations of the baselines and compare them with their LF-enhanced variants.
 Additional implementation details and a computational analysis (\ie inference speed and parameter count) are provided in the supplementary material.

\subsection{Main Results}
\input{paper/tables/main_results_NEW}
\input{paper/tables/framework_ablation}
\input{paper/tables/sd_results}
We evaluate lane segment and centerline perception, both with lane-lane topology. The former defines relations between full lane segments and is our primary setting, as it jointly captures lane geometry, including centerlines and boundaries, together with pedestrian crossings, while the latter enables comparison with a broader range of methods.
\\\noindent\textbf{Lane Segment perception.}
We conduct our main evaluation on OpenLane-V2 subset A~\cite{wang2023openlane}, using LaneSegNet~\cite{li2023lanesegnet} as the reference baseline. 
To isolate the contribution of the proposed LF framework, we keep the original LaneSegNet configuration unchanged and only integrate our modules. 
The model follows a camera-only setup with seven surround-view cameras, a ResNet-50 backbone, and a \mbox{BEVFormer} encoder producing a $100 \times 200$ BEV representation over a $\pm 50 \text{m} \times \pm 25 \text{m}$ perception range. 
We train for 24 epochs with a global batch size of 8, using AdamW and the original learning-rate schedule. 

Integrating our framework consistently improves LaneSegNet across all lane segment perception and topology metrics. As shown in \Cref{tab:openlanev2_joint_results}, our method achieves state-of-the-art performance on lane segment perception while remaining competitive on topology reasoning. This difference is expected, as the LF encodes local lane geometry rather than explicit topological relations. As a result, topology reasoning improves only indirectly through more accurate lane segment predictions.
\\\noindent\textbf{Centerline perception.}
To enable comparison with a broader range of methods, we additionally evaluate our framework in a centerline-based setting, considering only centerline perception and lane-lane associations between centerlines. We evaluate LaneSegNet under the same training protocol used for lane-segment perception, extracting centerlines from the predicted lane segments following~\cite{li2023lanesegnet}. We additionally apply our framework to TopoPoint~\cite{fu2026topopoint}, a native centerline perception and topology reasoning model. As shown in \Cref{tab:openlanev2_joint_results}, the gains are larger for LaneSegNet, as its lane-segment representation explicitly models the lane geometry targeted by our LF, whereas TopoPoint operates only on centerlines.

\subsection{Ablation on Lane Field Framework}
To evaluate the contribution of the proposed BEV feature modulation and deformable attention guidance modules, we perform a component-wise ablation by enabling one module at a time. 
\Cref{tab:framework_ablation} shows that each module contributes to the final performance, while their combination provides the strongest results.

\begin{figure*}[!t]
    \centering
    \includegraphics[width=\textwidth]{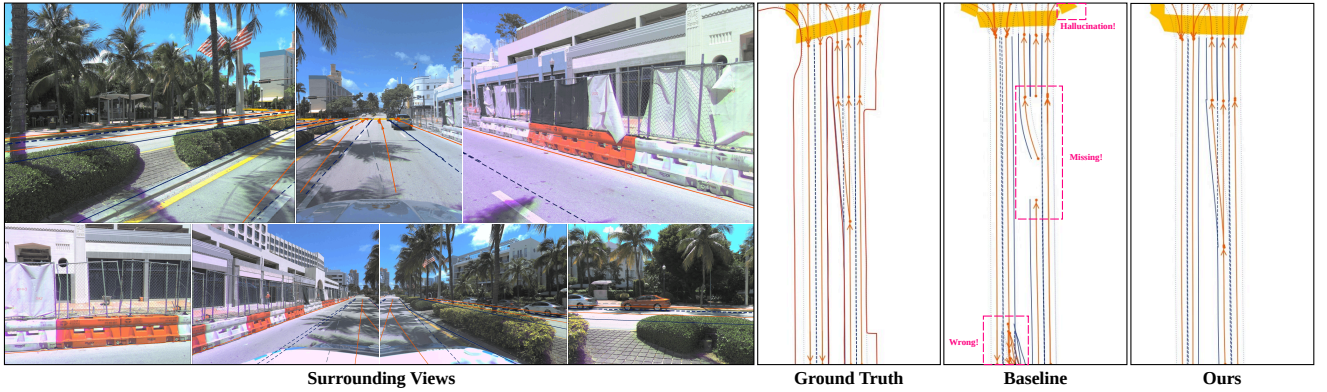}
\caption{
Qualitative results on OpenLane-V2.
We show the predicted map reprojected onto the perspective-view images and the corresponding BEV comparison among the ground truth, LaneSegNet baseline, and our Lane Field enhanced variant.
}
    \label{fig:HD_results}
\end{figure*}

\subsection{Ablation on SD Maps}
We evaluate our framework in a prior-based setting using the SD maps included in OpenLane-V2. We remove the field prediction head and derive the LF directly from their road polylines using the procedure in~\Cref{sec:road_field}. As they lack lane-width information, we use a fixed \(\sigma\) in the Gaussian line integral of~\Cref{eq:integral}.

 In this setting, the available SD maps are highly sparse and only provide road-level polylines, without lane count, lane width, or detailed semantic attributes. Richer map priors could therefore be used to construct more informative LFs, potentially improving performances. The SD-derived LF is then used as input to the BEV feature modulation block and the deformable attention guidance mechanism, as described in \Cref{sec:modulation} and~\Cref{sec:attention}. 
We perform this ablation on LaneSegNet~\cite{li2023lanesegnet} with OpenLane-V2, using the same setting as in~\Cref{tab:openlanev2_results}.
As reported in~\Cref{tab:sd_ablation}, the SD variant consistently improves LaneSegNet across all detection and topology metrics. 
This highlights again the flexibility of our framework: the LF can be either learned from BEV features or constructed from external map priors, such as SD maps, while still providing effective guidance for map construction. 
\subsection{Evaluation on Additional Datasets and Tasks}
\input{paper/tables/cross_datasets}
To assess the portability of our framework beyond lane segment perception and topology reasoning, we integrate it into MapTRv2~\cite{liao2025maptrv2} and GeMap \cite{zhang2024online} for map element detection on Argoverse~2~\cite{wilson2023argoverse} and NuScenes~\cite{caesar2020nuscenes}. We evaluate road boundaries, lane dividers, centerlines, and pedestrian crossings, which are represented as fragmented polylines rather than continuous lane segments. For each dataset, we construct the ground truth LF from the original vector annotations while retaining the baseline configuration. We train MapTRv2 for 6 epochs and GeMap for 12 epochs on Argoverse 2, and MapTRv2 for 24 epochs on NuScenes, using a batch size of 16 and a BEV range of $\pm 30\mathrm{m}\times\pm 15\mathrm{m}$. As shown in \Cref{tab:cross_datasets_ablation}, the largest gains are obtained for road boundaries and dividers, whose continuous and elongated geometry is well captured by our LF. 

\subsection{Qualitative Results}
\Cref{fig:HD_results} qualitatively compares LaneSegNet with our enhanced variant on OpenLane-V2. The baseline shows errors in both near and distant regions, including a hallucinated pedestrian crossing and less accurate map predictions. Our method better preserves the lane layout, producing a more accurate and structurally consistent HD map. Additional examples are provided in the supplementary material.

%% file: paper/tables/main_results_NEW.tex
\begin{table*}[t]
\centering
\small
\setlength{\tabcolsep}{3pt}
\renewcommand{\arraystretch}{1.12}

\begin{minipage}[t]{0.54\textwidth}
\vspace{0pt}
\centering
\caption*{\textbf{(a) Lane segment perception}}
\label{tab:openlanev2_results}

\begin{tabularx}{\linewidth}{@{}X c c c c@{}}
\toprule
\textbf{Method}
& \textbf{mAP}\;$\uparrow$
& \textbf{AP}$_{ls}\uparrow$
& \textbf{AP}$_{ped}\uparrow$
& \textbf{TOP}$_{lsls}\uparrow$ \\
\midrule

StreamMapNet \cite{yuan2024streammapnet} $^\ddagger$
& 20.3 & 22.1 & 18.6 & 13.2 \\

TopoNet \cite{li2026graph}
& 23.0 & 23.9 & 22.0 & -- \\

SQD-MapNet \cite{wang2024stream} $^\ddagger$
& 26.0 & 27.1 & 24.9 & 16.6 \\

MapTR \cite{liao2022maptr}
& 27.0 & 25.9 & 28.1 & -- \\

MapTRv2 \cite{liao2025maptrv2}
& 28.5 & 26.6 & 30.4 & -- \\

LaneSegNet \cite{li2023lanesegnet} $^\dagger$
& 32.1 & 32.2 & 31.9 & 25.5 \\

TopoLogic \cite{fu2024topologic}
& 33.2 & 33.0 & 33.4 & \textbf{30.8} \\

Topo2Seq \cite{yang2025topo2seq}
& \underline{33.6} & \underline{33.7} & \underline{33.5} & 26.9 \\

\midrule

\rowcolor{ResultGray}
\textbf{LaneSegNet + Ours}
& \textbf{36.1} {\scriptsize (+4.0)}
& \textbf{35.8} {\scriptsize (+3.6)}
& \textbf{36.4} {\scriptsize (+4.5)}
& \underline{28.4} {\scriptsize (+2.9)} \\

$\Delta_{\mathrm{rel}}$ vs. LaneSegNet
& {\scriptsize +12.5\%}
& {\scriptsize +11.2\%}
& {\scriptsize +14.1\%}
& {\scriptsize +11.4\%} \\

\bottomrule
\end{tabularx}
\end{minipage}
\hfill
\begin{minipage}[t]{0.44\textwidth}
\vspace{0pt}
\centering
\caption*{\textbf{(b) Centerline perception}}
\label{tab:openlanev2_centerline_results}

\begin{tabularx}{\linewidth}{@{}X c c c@{}}
\toprule
\textbf{Method}
& \textbf{DET}$_l\uparrow$
& \textbf{TOP}$_{ll}\uparrow$
& \textbf{OLS}$_l\uparrow$ \\
\midrule



MapTR \cite{liao2022maptr}
& 17.7 & 5.9 & 21.0 \\

TopoNet \cite{li2026graph}
& 28.6 & 10.9 & 30.8 \\

TopoMLP \cite{wu2024topomlp}
& 28.3 & 21.6 & 37.4 \\

TopoLogic \cite{fu2024topologic}
& 29.9 & 23.9 & 39.4 \\

LaneSegNet \cite{li2023lanesegnet} $^\dagger$
& 31.1 & 25.0 & 40.6 \\

TopoFormer \cite{lv2025t2sg} $^\ast$
& \underline{34.7} & 24.1 & 41.9 \\

TopoPoint \cite{fu2026topopoint} $^\dagger$
& 31.1 & \underline{28.5} & 42.2 \\

Topo2Seq \cite{yang2025topo2seq}
& 33.5 & 27.0 & 42.7 \\

\midrule

\rowcolor{ResultGray}
\textbf{LaneSegNet + Ours}
& \textbf{35.2} {\scriptsize (+4.1)}
& 28.0 {\scriptsize (+3.0)}
& \textbf{44.1} {\scriptsize (+3.5)} \\

\rowcolor{ResultGray}
\textbf{TopoPoint + Ours}
& 32.8 {\scriptsize (+1.7)}
& \textbf{30.0} {\scriptsize (+1.5)}
& \underline{43.8} {\scriptsize (+1.6)} \\

\bottomrule
\end{tabularx}
\end{minipage}

\caption{Comparison on OpenLane-V2 subset A. \textbf{(a)} Results on lane segment perception. \textbf{(b)} Results on centerline perception, where OLS$_{l}$ is computed using only centerline detection and lane-lane topology. All models adopt ResNet-50 as backbone and are trained for 24 epochs. $^\dagger$: Denotes the baseline trained from the official public implementation. $^\ddagger$: Denotes methods that exploit temporal information.
 $^\ast$: TopoFormer uses a pre-trained lane detector. The best and second best results are highlighted in \textbf{bold} and \underline{underlined}, respectively, while ``--'' indicates no relevant results. All the deltas are reported with respect to the corresponding baseline.}
\label{tab:openlanev2_joint_results}
\end{table*}

%% file: paper/tables/framework_ablation.tex
\begin{table}[t]
\centering
\small
\setlength{\tabcolsep}{4pt}
\renewcommand{\arraystretch}{1.15}

\newcommand{\res}[2]{#1 {\scriptsize(#2)}}
\newcommand{\bres}[2]{\textbf{#1} {\scriptsize{(#2)}}}

\begin{tabularx}{\columnwidth}{
l
>{\centering\arraybackslash}X
>{\centering\arraybackslash}X
>{\centering\arraybackslash}X
>{\centering\arraybackslash}X
}
\hline
\textbf{Component} & \textbf{mAP}\;$\uparrow$ & \textbf{AP$_{ls}$}\;$\uparrow$ & \textbf{AP$_{ped}$}\;$\uparrow$ & \textbf{TOP$_{lsls}$}\;$\uparrow$ \\
\hline
Baseline   
& 32.1 
& 32.2 
& 31.9
& 25.5 \\

+ Guidance  
& \res{33.2}{+1.1} 
& \res{33.4}{+1.2} 
& \res{32.9}{+1.0}
& \res{26.8}{+1.3} \\

+ Modulation        
& \res{34.9}{+2.8} 
& \res{35.7}{+3.5} 
& \res{34.0}{+2.1}
& \res{27.9}{+2.4} \\
\hline

\rowcolor{ResultGray}
\textbf{+ Both}              
& \bres{36.1}{+4.0} 
& \bres{35.8}{+3.6} 
& \bres{36.4}{+4.5}
& \bres{28.4}{+2.9} \\
\hline
\end{tabularx}

\caption{Component-wise ablation on OpenLane-V2 with LaneSegNet, evaluating BEV feature modulation, deformable attention guidance, and their combination.}
\label{tab:framework_ablation}
\end{table}

%% file: paper/tables/sd_results.tex
\begin{table}[t]
\centering
\small
\setlength{\tabcolsep}{4pt}
\renewcommand{\arraystretch}{1.15}

\newcommand{\bres}[2]{\textbf{#1} {\scriptsize(#2)}}

\begin{tabularx}{\columnwidth}{
l
>{\centering\arraybackslash}X
>{\centering\arraybackslash}X
>{\centering\arraybackslash}X
>{\centering\arraybackslash}X
}
\hline
\textbf{Method} 
& \textbf{mAP}\;$\uparrow$
& \textbf{AP$_{ls}$}\;$\uparrow$ 
& \textbf{AP$_{ped}$}\;$\uparrow$ 
& \textbf{TOP$_{lsls}$}\;$\uparrow$ \\
\hline

LaneSegNet
& 32.1 
& 32.2 
& 31.9 
& 25.5 \\

\rowcolor{ResultGray}
\textbf{+ Ours (SD)}
& \bres{36.8}{+4.7} 
& \bres{36.6}{+4.4} 
& \bres{37.0}{+5.1} 
& \bres{29.7}{+4.2} \\

$\Delta_{\mathrm{rel}}$
& {\scriptsize +14.6\%}
& {\scriptsize +13.7\%}
& {\scriptsize +16.0\%}
& {\scriptsize +16.5\%} \\
\hline
\end{tabularx}

\caption{
Comparison between the LaneSegNet baseline and the variant using SD maps derived lane fields on OpenLane-V2.
}
\label{tab:sd_ablation}
\end{table}

%% file: paper/tables/cross_datasets.tex
\begin{table}[t]
\centering
\small
\setlength{\tabcolsep}{2.2pt}
\renewcommand{\arraystretch}{1.12}

\begin{tabularx}{\columnwidth}{
@{}
l
l
>{\centering\arraybackslash}X
>{\centering\arraybackslash}X
>{\centering\arraybackslash}X
>{\centering\arraybackslash}X
>{\centering\arraybackslash}X
@{}
}
\hline
\textbf{Dataset} & \textbf{Method} 
& \textbf{mAP}\;$\uparrow$ 
& \textbf{AP$_{div}$}$\uparrow$ 
& \textbf{AP$_{ped}$}$\uparrow$ 
& \textbf{AP$_{bou}$}$\uparrow$ 
& \textbf{AP$_{cen}$}$\uparrow$ \\
\hline

\multirow{4}{*}{AV2} 
& MapTRv2
& 62.0 
& 66.3 
& 54.8 
& 64.6 
& 62.0 \\

& \cellcolor{ResultGray}\textbf{+ Ours}
& \cellcolor{ResultGray}\textbf{64.2}
& \cellcolor{ResultGray}\textbf{69.5}
& \cellcolor{ResultGray}\textbf{56.4}
& \cellcolor{ResultGray}\textbf{67.1}
& \cellcolor{ResultGray}\textbf{63.6} \\

\cline{2-7}

& GeMap
& 70.3 
& 74.6
& 66.7 
& 69.5 
& -- \\

& \cellcolor{ResultGray}\textbf{+ Ours}
& \cellcolor{ResultGray}\textbf{71.2}
& \cellcolor{ResultGray}\textbf{75.3}
& \cellcolor{ResultGray}\textbf{67.4}
& \cellcolor{ResultGray}\textbf{70.8}
& \cellcolor{ResultGray}\textbf{--} \\
\hline

\multirow{2}{*}{NUS} 
& MapTRv2
& 55.8 
& 55.9 
& 52.5 
& 60.1
& 54.9 \\

& \cellcolor{ResultGray}\textbf{+ Ours}
& \cellcolor{ResultGray}\textbf{57.7}
& \cellcolor{ResultGray}\textbf{59.2}
& \cellcolor{ResultGray}\textbf{52.9}
& \cellcolor{ResultGray}\textbf{62.8}
& \cellcolor{ResultGray}\textbf{55.8} \\
\hline

\end{tabularx}

\caption{
Analysis on Argoverse 2 and NuScenes using MapTRv2 and GeMap as baselines for the standard map element detection.
}
\label{tab:cross_datasets_ablation}
\end{table}

%% file: paper/sec/5_conclusion.tex
\section{Conclusion}
We introduced the Lane Field, a dense BEV-aligned representation encoding local lane geometry through direction, proximity, and confidence. It integrates into online HD mapping architectures through two plug-and-play modules: BEV feature modulation, which enriches latent BEV features, and deformable attention guidance, which steers sampling according to local lane geometry.

The LF supports both learned and prior-based settings. In the learned setting, it is predicted under supervision from ground truth fields analytically derived from map annotations. In the prior-based setting, it is constructed from external SD maps to exploit available map priors. Experiments confirm that our framework generalizes across datasets and baselines, consistently improving performances.

We acknowledge that our current method is inherently lane-centric and does not explicitly represent elements such as pedestrian crossings. Future works could extend the LF to model crosswalks and other map structures. Moreover, while BEV feature modulation is broadly applicable, attention guidance is currently limited to deformable attention, motivating future integrations to widen LF compatibility.

%% file: supplementary/sec/0_overview.tex
\section*{Overview of Supplementary Material}
\noindent The supplementary material contains these parts:
\begin{enumerate}[label=\Alph*.]
    \item Mathematical Derivation of Field Weights $I_i(p)$.
    \item Analysis of Gaussian Spread $\sigma$.
    \item Additional implementation details.
    \item Mathematical Definition of Loss Functions.
    \item Qualitative Results on Field Variance.
    \item Analysis of Guided Deformable Attention.
    \item Analysis of BEV Features Modulation.
    \item Additional Qualitative Results.
    \item Inference Speed and Model Size.
    \item Ablation on Gradient Flow.
\end{enumerate}

%% file: supplementary/sec/1_derivation.tex
\section{Mathematical Derivation of Field Weights}
We map sparse vectorized lane geometry onto the dense BEV grid by defining each
segment contribution as the line integral of a Gaussian distance kernel along
the segment.
Unlike closest-point methods, our formulation integrates over the entire segment, ensuring smooth contributions.
Let segment \(i\) be defined by its endpoints \(A_i\) and \(B_i\). We
parametrize any point along the segment as
\begin{equation}
q_i(t) = A_i + t(B_i-A_i), \qquad t\in[0,1],
\end{equation}
where \(q_i(0)=A_i\) and \(q_i(1)=B_i\). Thus, \(t\) is a normalized coordinate
along the segment, and \(q_i(t)\) represents a point moving continuously from
\(A_i\) to \(B_i\). For a BEV location \(p\), the vector \(p-q_i(t)\) connects
\(p\) to the point \(q_i(t)\) on the segment, and \(\|p-q_i(t)\|^2\) is their
squared Euclidean distance.
The contribution of segment \(i\) to location \(p\) is obtained by integrating:
\begin{equation}
I_i(p)
=
\ell_i
\int_0^1
\exp\left(
-\frac{\|p - A_i - t(B_i-A_i)\|^2}{2\sigma_i^2}
\right) dt,
\end{equation}
where \(\ell_i\) is the segment length. The Gaussian term gives higher weight
to segment points closer to \(p\) and smoothly decreases the contribution as
their distance increases.

To derive the closed-form solution, we first rewrite the squared distance
between the BEV location \(p\) and the points along the segment. The key observation is that the distance between \(p\) and a point on
the segment can be decomposed into a perpendicular component to the line
supporting the segment and a longitudinal component along the segment:
\begin{equation}
\|p-A_i-t(B_i-A_i)\|^2
=
d_{\perp,i}^2 + \ell_i^2(t-t_{0,i})^2,
\end{equation}
where \(d_{\perp,i}\) is the perpendicular distance from \(p\) to the supporting
line, and \(t_{0,i}\) is the normalized coordinate of the orthogonal projection
of \(p\) onto this line.

\noindent Let \(\mathbf{d}_i=B_i-A_i\). The projection coordinate is given by
\begin{equation}
t_{0,i}
=
\frac{(p-A_i)^\top \mathbf{d}_i}{\ell_i^2}.
\end{equation}
The corresponding perpendicular distance is
\begin{equation}
d_{\perp,i}^2
=
\|p-A_i\|^2
-
\frac{\left((p-A_i)^\top \mathbf{d}_i\right)^2}{\ell_i^2}.
\end{equation}
Note that, \(t_{0,i}\) is not clamped to \([0,1]\): it denotes the projection
onto the infinite supporting line, while the finite segment endpoints are
handled by the integration bounds.

\noindent Using the distance decomposition yields
\begin{equation}
I_i(p)
=
\ell_i
\int_0^1
\exp\left(
-\frac{
d_{\perp,i}^2 + \ell_i^2(t-t_{0,i})^2
}{2\sigma_i^2}
\right)dt.
\end{equation}
Since the perpendicular distance \(d_{\perp,i}\) does not depend on \(t\), the
corresponding term is constant along the integration domain and can be factored
out:
\begin{equation}
I_i(p)
=
\ell_i
\exp\left(
-\frac{d_{\perp,i}^2}{2\sigma_i^2}
\right)
\int_0^1
\exp\left(
-\frac{\ell_i^2(t-t_{0,i})^2}{2\sigma_i^2}
\right)dt.
\end{equation}

\noindent The remaining term is a one-dimensional Gaussian integral along the segment
coordinate. To evaluate it, we use the change of variable
\begin{equation}
u =
\frac{\ell_i(t-t_{0,i})}{\sqrt{2}\sigma_i},
\qquad
dt =
\frac{\sqrt{2}\sigma_i}{\ell_i}du.
\end{equation}
The integration bounds become
\begin{equation}
u_{0,i} =
\frac{-\ell_i t_{0,i}}{\sqrt{2}\sigma_i},
\qquad
u_{1,i} =
\frac{\ell_i(1-t_{0,i})}{\sqrt{2}\sigma_i}.
\end{equation}
Therefore,
\begin{equation}
\int_0^1
\exp\left(
-\frac{\ell_i^2(t-t_{0,i})^2}{2\sigma_i^2}
\right)dt
=
\frac{\sqrt{2}\sigma_i}{\ell_i}
\int_{u_{0,i}}^{u_{1,i}} e^{-u^2}du.
\end{equation}

\noindent The Gaussian integral \(\int e^{-u^2}du\) has no elementary indefinite form and
is expressed through the error function:
\begin{equation}
\operatorname{erf}(u)
=
\frac{2}{\sqrt{\pi}}
\int_0^u e^{-s^2}ds.
\end{equation}
Hence,
\begin{equation}
\int_{u_{0,i}}^{u_{1,i}} e^{-u^2}du
=
\frac{\sqrt{\pi}}{2}
\left[
\operatorname{erf}(u_{1,i}) - \operatorname{erf}(u_{0,i})
\right].
\end{equation}
Substituting this result back into \(I_i(p)\), and simplifying the factor
\(\ell_i\), yields the closed-form segment weight:
\begin{equation}
I_i(p)
=
\frac{\sigma_i\sqrt{2\pi}}{2}
\exp\left(
-\frac{d_{\perp,i}^2}{2\sigma_i^2}
\right)
\left[
\operatorname{erf}(u_{1,i}) - \operatorname{erf}(u_{0,i})
\right].
\end{equation}

%% file: supplementary/sec/2_sigma_ablation.tex
\section{Analysis of Gaussian Spread }
\begin{figure*}[!t]
    \centering

    \includegraphics[width=0.24\textwidth]{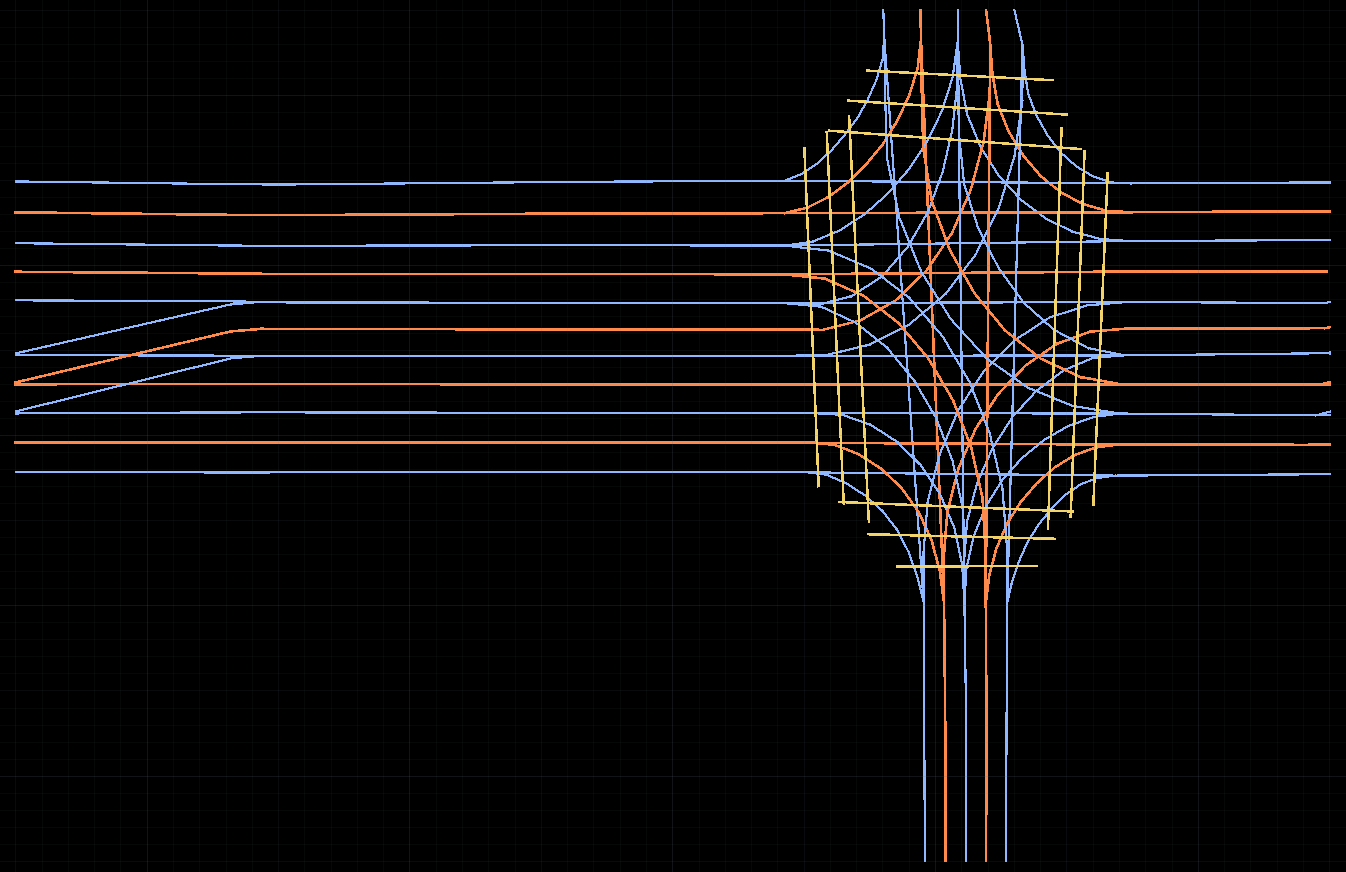}
    \hfill
    \includegraphics[width=0.24\textwidth]{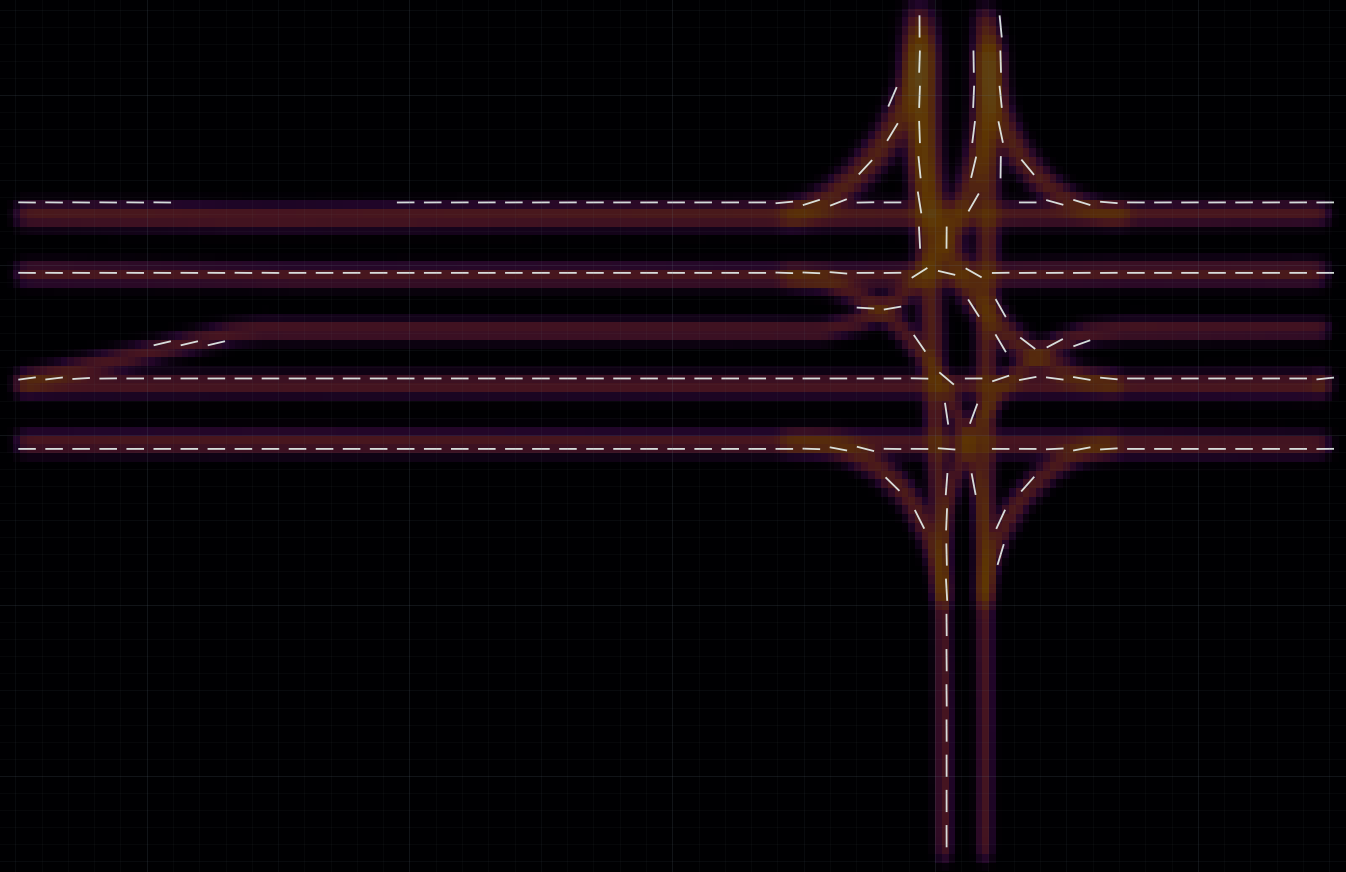}
    \hfill
    \includegraphics[width=0.24\textwidth]{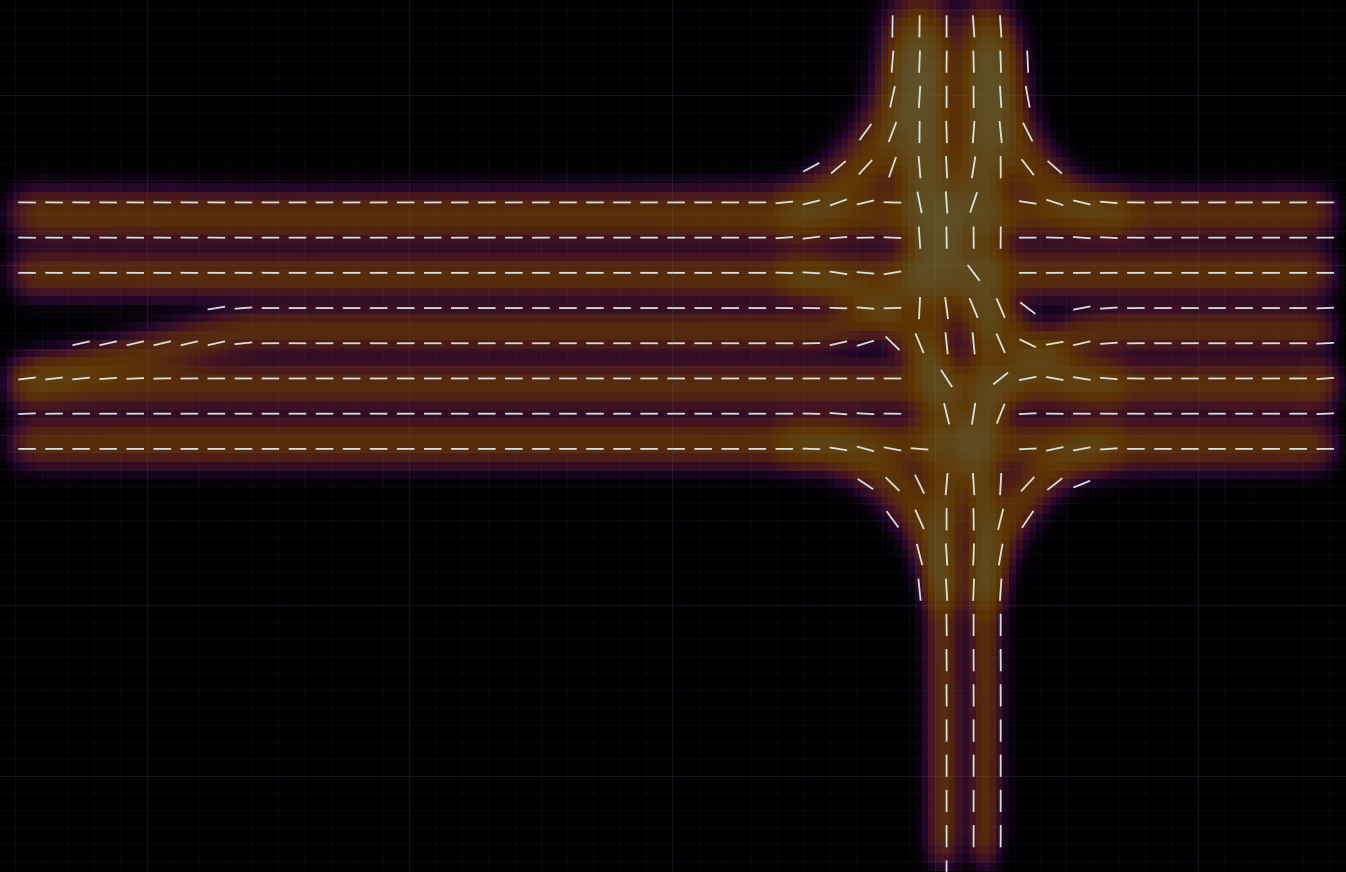}
    \hfill
    \includegraphics[width=0.24\textwidth]{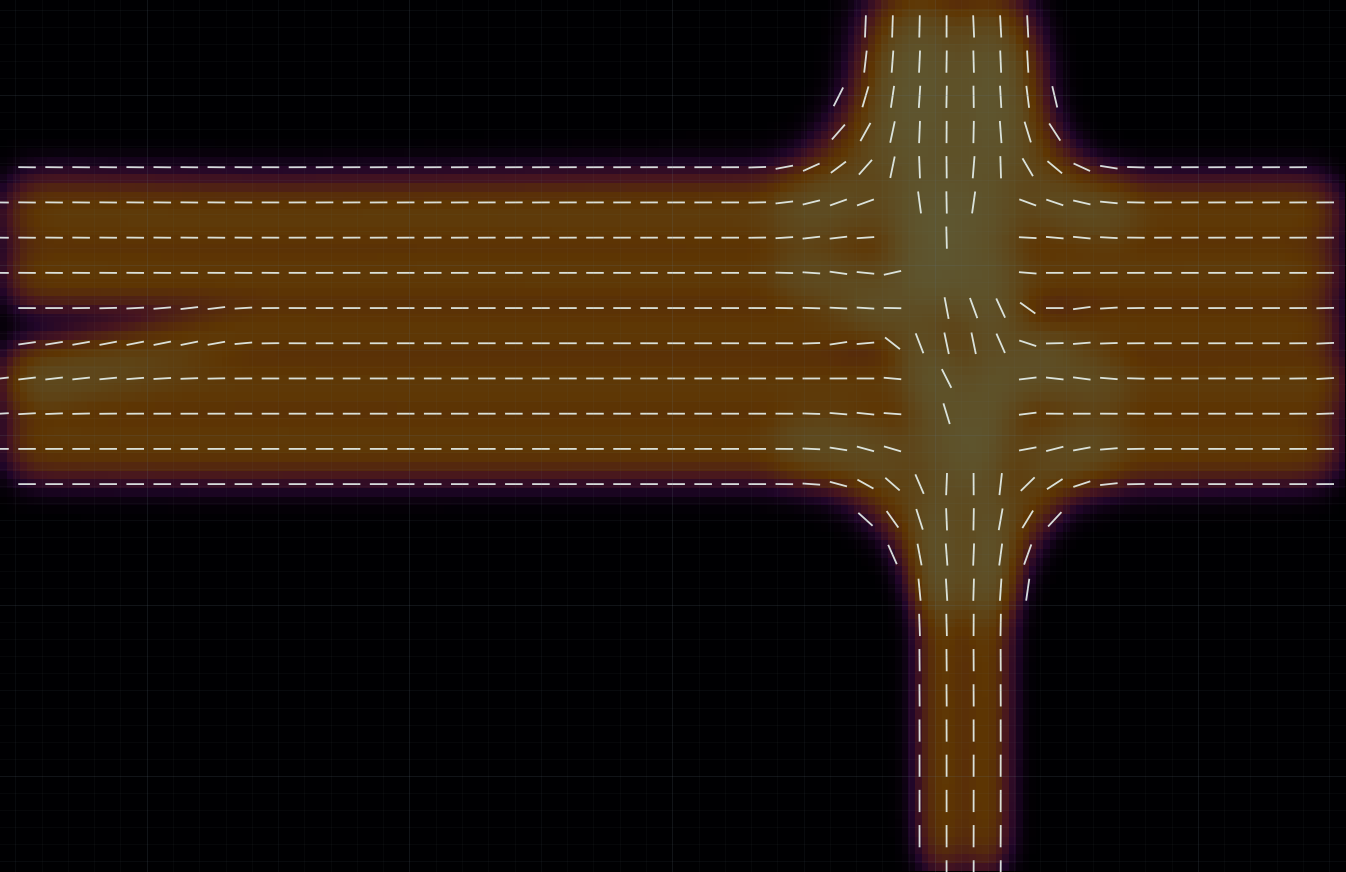}

\caption{
Effect of the spread factor \(f\) on the generated ground truth Lane Field. From left to right, fields are generated with \(f=0.25\), \(f=0.50\), and \(f=0.75\).
}
    \label{fig:sigma_qualitative}
\end{figure*}
The spread parameter \(\sigma_i\) controls the spatial extent of the contribution
of segment \(i\) over the BEV grid. Rather than using a fixed value for all
segments, we adapt \(\sigma_i\) to the local lane width as
\begin{equation}
\sigma_i = \max(\bar{w}_i\cdot f, \sigma_{\min}),
\end{equation}
where \(f\) is a scale factor and \(\sigma_{\min}\) prevents degenerate values
for very narrow or noisy lane annotations. The factor \(f\) determines how much
the Gaussian support expands around the centerline: small values produce sharp
and localized fields, while larger values generate broader and smoother fields. To interpret the role of \(f\), consider the Gaussian value at the lane boundary,
\ie at distance \(\bar{w}_i/2\) from the centerline. Ignoring the longitudinal
term, this value is
\begin{equation}
\exp\left(
-\frac{(\bar{w}_i/2)^2}{2(\bar{w}_i f)^2}
\right)
=
\exp\left(
-\frac{1}{8f^2}
\right).
\end{equation}
Thus, \(f\) directly controls how much signal is preserved at the lane boundary.
A small \(f\) rapidly suppresses the field away from the centerline, whereas a
large \(f\) keeps high responses even near and beyond the lane boundaries.
We choose \(f\) based on the qualitative analysis in \Cref{fig:sigma_qualitative} and \Cref{tab:sigma_factor}, finding that \(f=0.5\) provides the most balanced field.
\input{supplementary/tables/sigma_ablation}

%% file: supplementary/tables/sigma_ablation.tex

\begin{table}[t]
\centering
\small
\setlength{\tabcolsep}{4pt}
\renewcommand{\arraystretch}{1.05}

\begin{tabular*}{\columnwidth}{@{\extracolsep{\fill}} c c c}
\toprule
\(f\) & \(\sigma_i\) & Weight at lane edge \\
\midrule
0.25 & \(\bar{w}_i/4\)        & \(13\%\) \\
0.50 & \(\bar{w}_i/2\)        & \(61\%\) \\
0.75 & \(0.75\bar{w}_i\)      & \(80\%\) \\
1.00 & \(\bar{w}_i\)          & \(88\%\) \\
\bottomrule
\end{tabular*}

\caption{
Effect of the scale factor \(f\) on the Gaussian weight at the lane edge, i.e., at distance \(\bar{w}_i/2\) from the centerline.
}
\label{tab:sigma_factor}
\end{table}

%% file: supplementary/sec/2.5_implementation_details.tex
\section{Additional Implementation Details}
\label{sec:additional_impl}

This section reports implementation details omitted from the main paper
for clarity, but useful to fully specify the ground truth Lane Field
target used for supervision. In practice, we use a saturated version of
the proximity accumulation:
\begin{equation}
\mathbf{P}^{\mathrm{gt}}(p)
=
\frac{\mathbf{P}(p)}
{\mathbf{P}(p)+\tau_P}
=
\frac{\sum_i I_i(p)}
{\sum_i I_i(p)+\tau_P},
\end{equation}
with \(\tau_P=1.0\). This bounds the proximity target in \([0,1]\) while
preserving relative differences in low-evidence regions. As briefly mentioned in the main paper, the directional tensor is
trace-normalized and softly masked by proximity:
\begin{equation}
\mathbf{M}^{\mathrm{gt}}(p)
=
\frac{\mathbf{M}(p)}
{\operatorname{tr}(\mathbf{M}(p))+\varepsilon}
\cdot
\sigma\!\left(k\cdot(\mathbf{P}^{\mathrm{gt}}(p)-\tau_m)\right).
\end{equation}
The same proximity mask is applied to the direction confidence channel:
\begin{equation}
\mathbf{C}^{\mathrm{gt}}(p)
=
\mathbf{C}(p)
\cdot
\sigma\!\left(k\cdot(\mathbf{P}^{\mathrm{gt}}(p)-\tau_m)\right).
\end{equation}
We use \(k=20\) and \(\tau_m=0.3\). No additional normalization is applied
to \(\mathbf{C}^{\mathrm{gt}}(p)\), since it is derived from the anisotropy
of the normalized directional tensor.

%% file: supplementary/sec/3_loss_functions.tex
\section{Mathematical Definition of Loss Functions}
\begin{figure*}[t]
    \centering

    \begin{minipage}{0.32\textwidth}
        \centering
        \includegraphics[width=\linewidth]{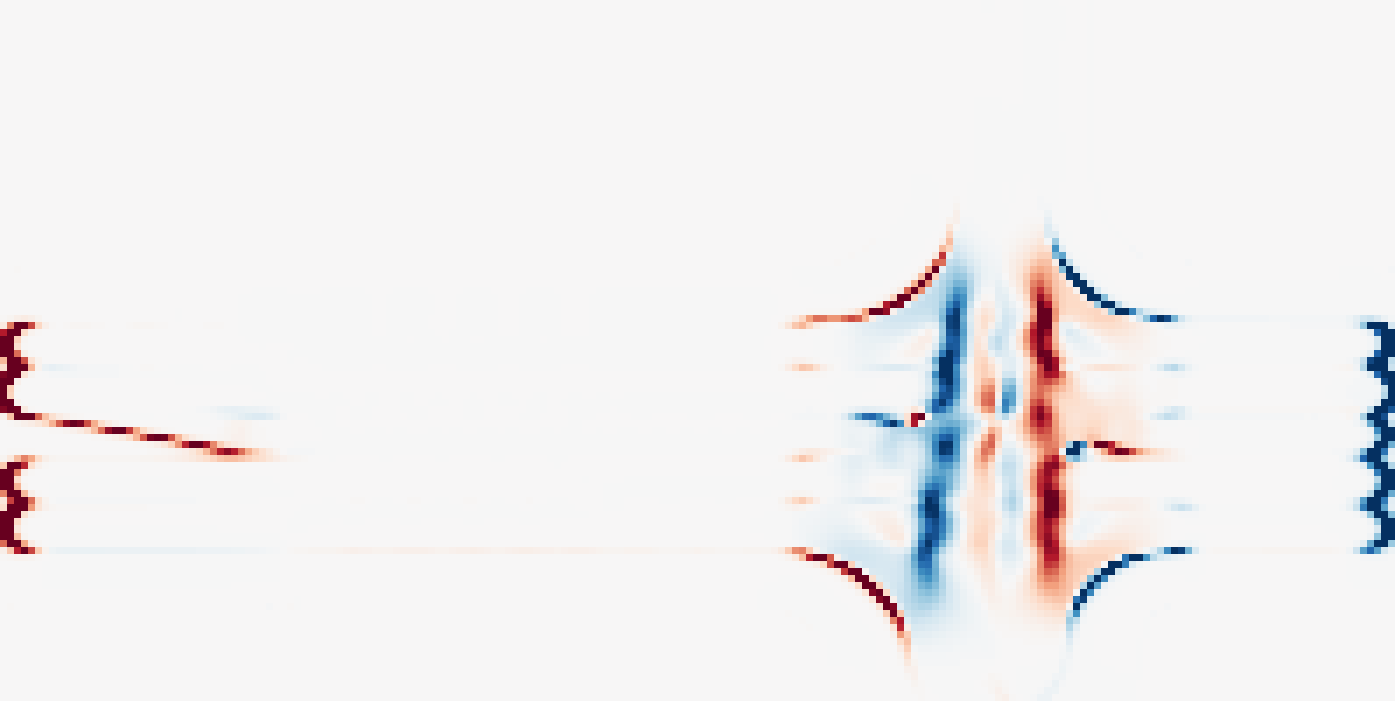}
    \end{minipage}
    \hfill
    \begin{minipage}{0.32\textwidth}
        \centering
        \includegraphics[width=\linewidth]{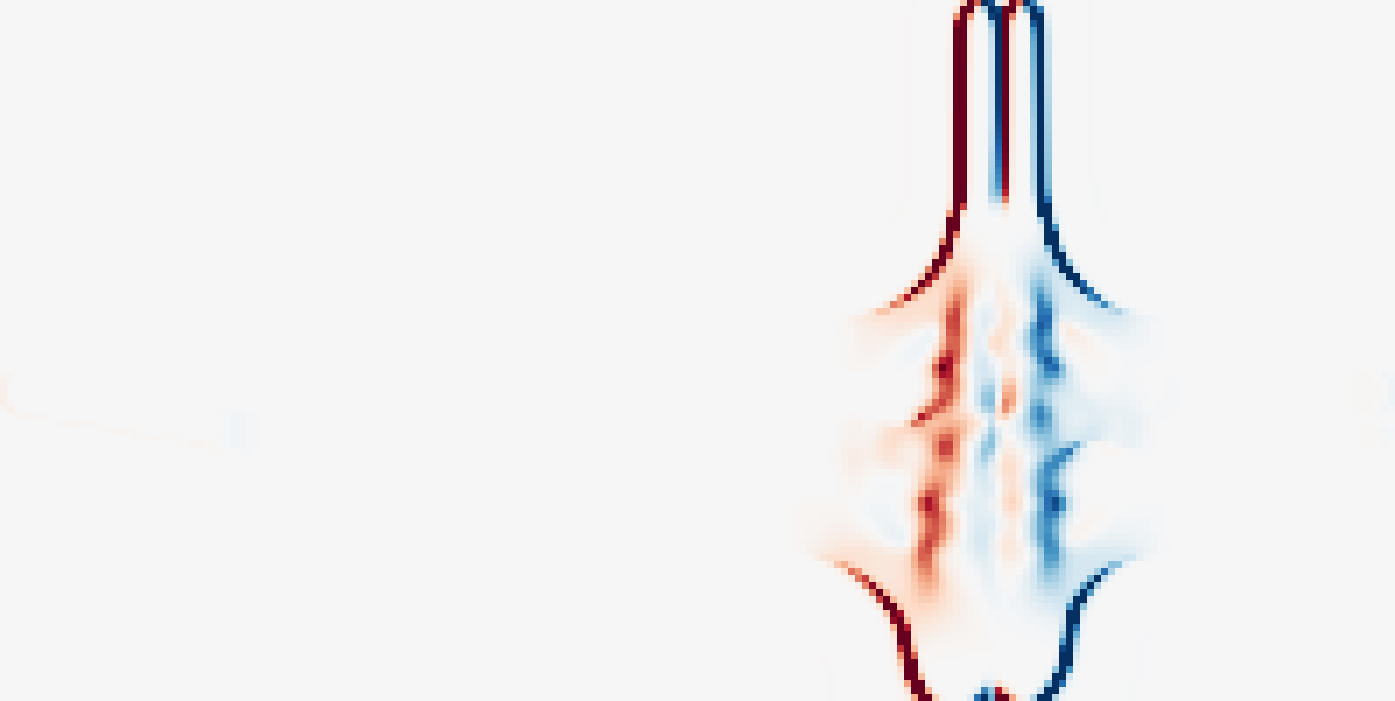}
    \end{minipage}
    \hfill
    \begin{minipage}{0.32\textwidth}
        \centering
        \includegraphics[width=\linewidth]{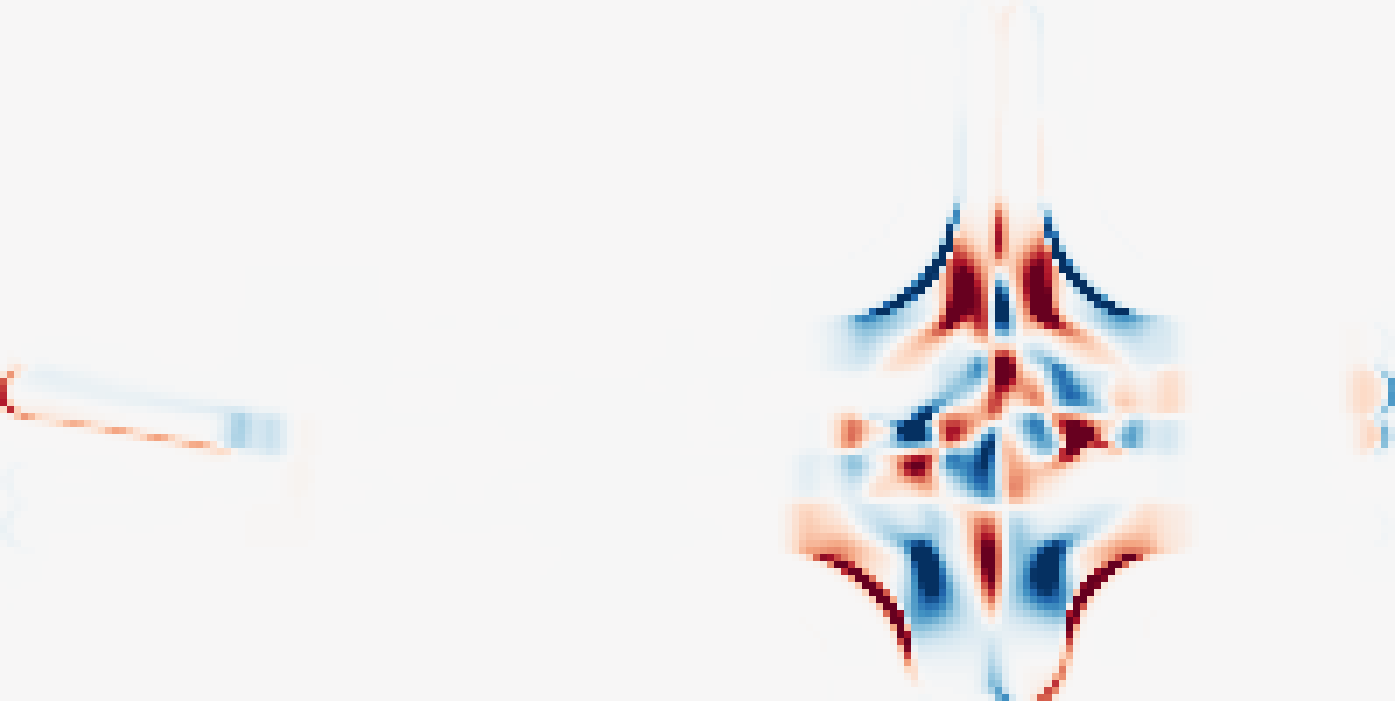}
    \end{minipage}

    \vspace{0.5em}

    \begin{minipage}{0.32\textwidth}
        \centering
        \includegraphics[width=\linewidth]{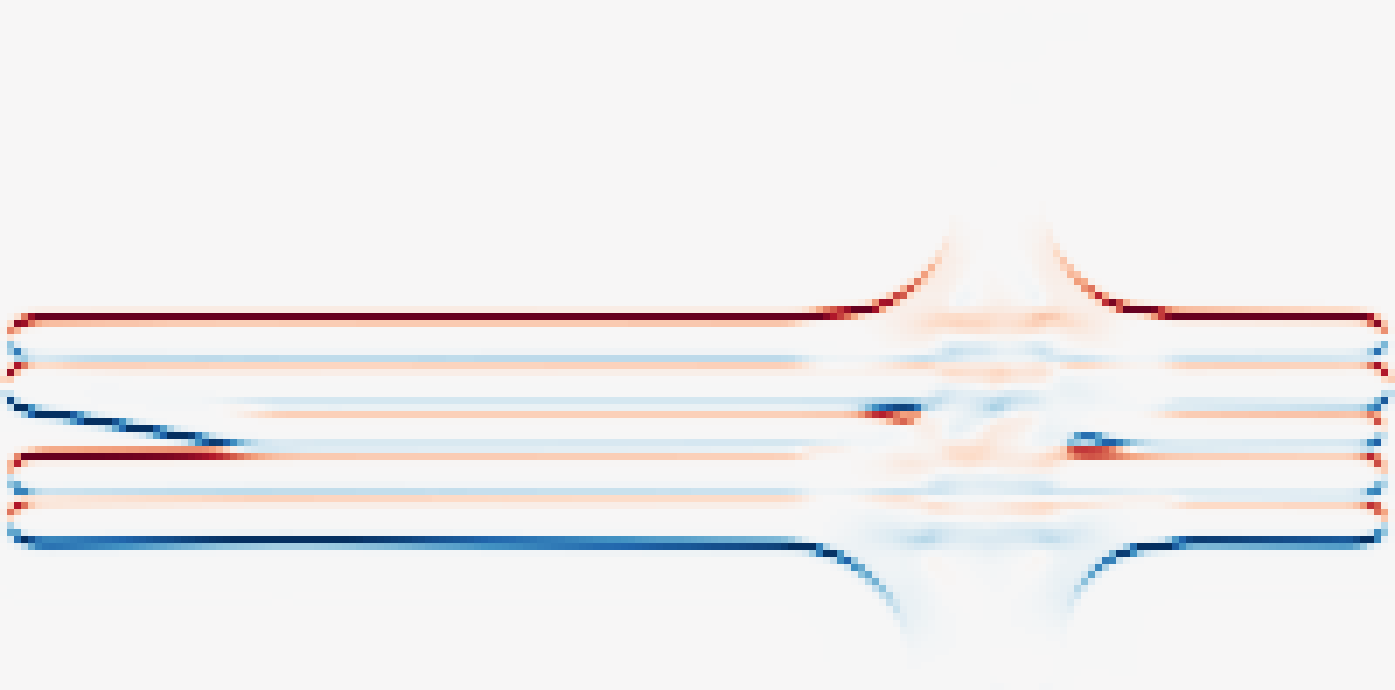}
    \end{minipage}
    \hfill
    \begin{minipage}{0.32\textwidth}
        \centering
        \includegraphics[width=\linewidth]{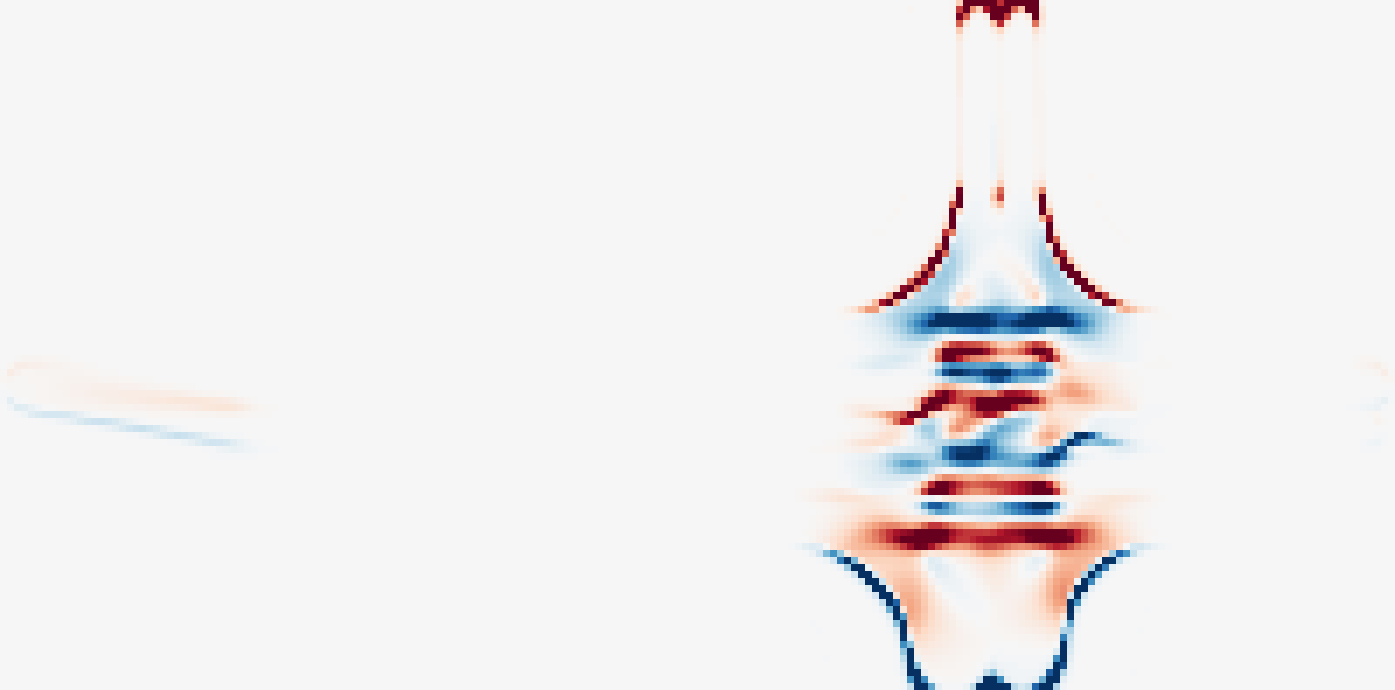}
    \end{minipage}
    \hfill
    \begin{minipage}{0.32\textwidth}
        \centering
        \includegraphics[width=\linewidth]{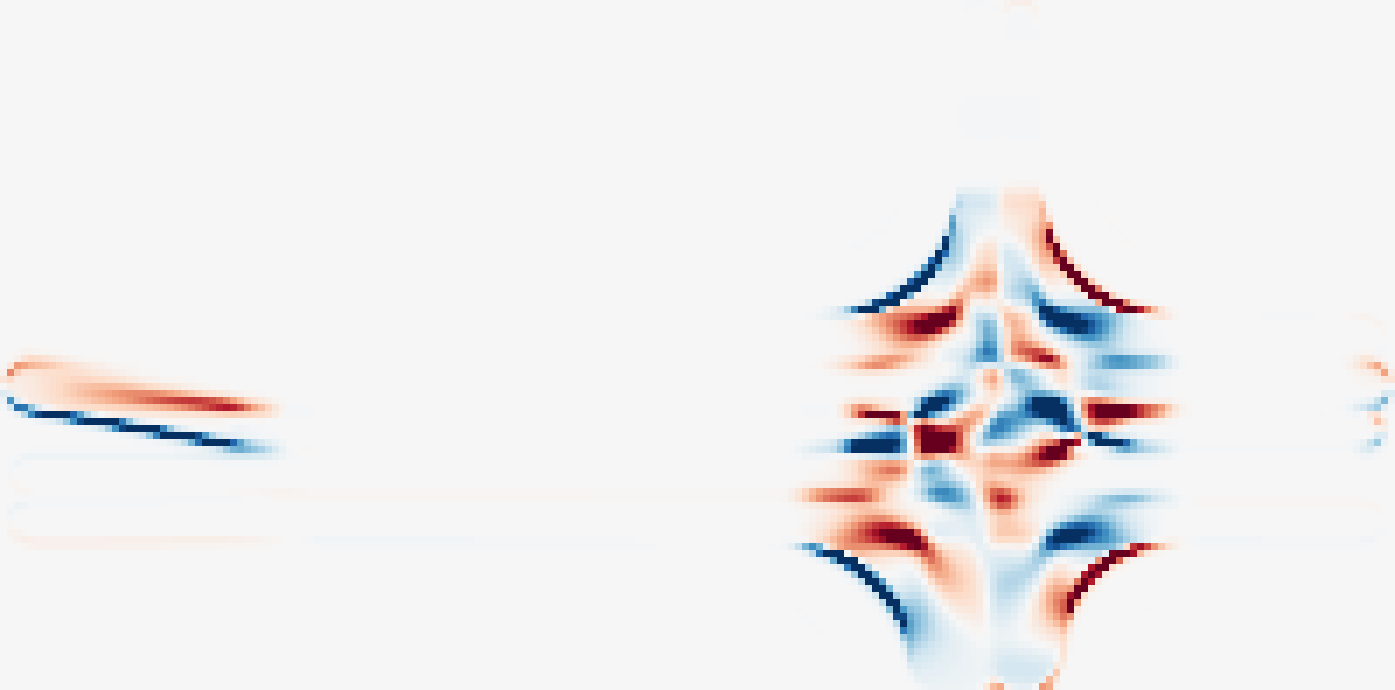}
    \end{minipage}

    \caption{Visualization of the spatial gradients of the ground truth tensor channels along the horizontal ($x$) and vertical ($y$) directions. The first row shows $\Delta_x$ for $M_{xx}$, $M_{yy}$, and $M_{xy}$, while the second row shows $\Delta_y$ for the same channels.}
    \label{fig:road_field_gradients}
\end{figure*}

We optimize the LF prediction head using three complementary losses. 
The total objective is defined as a weighted sum of a regression loss, a spatial gradient loss, and a proximity uncertainty loss:
\begin{equation}
    \mathcal{L}
    =
    w_R \cdot \mathcal{L}_{R}
    +
    w_G \cdot \mathcal{L}_{G}
    +
    w_L \cdot \mathcal{L}_{L},
\end{equation}
where \(w_R\), \(w_G\), and \(w_L\) control the relative contribution of each term.
\\\\
\noindent\textbf{Regression Loss.}
The regression loss supervises the directional tensor channels of the LF.
Let \(\mathbf{M}(p)\) and \(\hat{\mathbf{M}}(p)\) denote the ground truth and predicted directional channels at BEV location \(p\), respectively.
We apply a Smooth-\(L_1\) loss \cite{girshick2015fast} at each BEV location and aggregate it with mean reduction:
\begin{equation}
\mathcal{L}_{R}
=
\frac{1}{3|\Omega|}
\sum_{p \in \Omega}
\sum_{c \in \{xx,xy,yy\}}
\ell_{\mathrm{smooth}}
\left(
\hat{\mathbf{M}}_{c}(p),
\mathbf{M}_{c}(p)
\right),
\end{equation}
where \(\Omega\) denotes the set of BEV locations, and the Smooth-\(L_1\) loss is defined as
\begin{equation}
\ell_{\mathrm{smooth}}(\hat{x},x)
=
\begin{cases}
\frac{1}{2}(\hat{x}-x)^2, & \text{if } |\hat{x}-x| < 1,\\
|\hat{x}-x|-\frac{1}{2}, & \text{otherwise}.
\end{cases}
\end{equation}
\\\\
\noindent\textbf{Gradient Regression Loss.}
Inspired by the Gradient Difference Loss of Mathieu \etal \cite{mathieu2015deep}, we introduce an additional \(L_1\) loss on the spatial gradients of the directional tensor channels.
For each channel \(\mathbf{M}_{xx}\), \(\mathbf{M}_{xy}\), and \(\mathbf{M}_{yy}\), 
we approximate spatial gradients with first-order finite differences along the \(x\) and \(y\) directions, 
and penalize the discrepancy between predicted and ground truth gradients.
As shown in \Cref{fig:road_field_gradients}, this term encourages sharper geometric edges and better preserves local transitions in the LF. 

\begin{equation}
\Delta_x \mathbf{M}_c(i,j)
=
\mathbf{M}_c(i,j+1)-\mathbf{M}_c(i,j),
\end{equation}

\begin{equation}
\Delta_y \mathbf{M}_c(i,j)
=
\mathbf{M}_c(i+1,j)-\mathbf{M}_c(i,j).
\end{equation}

\begin{equation}
\mathcal{L}_{G}^{x}
=
\frac{1}{3|\Omega_x|}
\sum_{p \in \Omega_x}
\sum_{c \in \{xx,xy,yy\}}
\left|
\Delta_x \hat{\mathbf{M}}_{c}(p)
-
\Delta_x \mathbf{M}_{c}(p)
\right|,
\end{equation}

\begin{equation}
\mathcal{L}_{G}^{y}
=
\frac{1}{3|\Omega_y|}
\sum_{p \in \Omega_y}
\sum_{c \in \{xx,xy,yy\}}
\left|
\Delta_y \hat{\mathbf{M}}_{c}(p)
-
\Delta_y \mathbf{M}_{c}(p)
\right|,
\end{equation}

\begin{equation}
\mathcal{L}_{G}
=
\mathcal{L}_{G}^{x}
+
\mathcal{L}_{G}^{y}.
\end{equation}
where \(\Omega_x\) and \(\Omega_y\) denote the valid BEV locations for horizontal and vertical finite differences, respectively. Differently from \cite{mathieu2015deep}, we compare signed finite differences rather than their magnitudes, since the direction of local variations is relevant for our field.
\\\\
\noindent\textbf{Laplacian \(\beta\)-NLL Loss.}
To supervise the proximity channel \(\mathbf{P}(p)\), we employ a Laplacian \(\beta\)-NLL loss~\cite{gu2024producing}, where the head predicts both the proximity mean and the corresponding log-scale. 
This can be interpreted as predicting a Laplacian distribution at each BEV location, whose scale encodes the uncertainty of the proximity estimate. 
Beyond regressing the proximity field, this uncertainty provides an explicit reliability cue for the BEV feature modulation block and the deformable attention guidance mechanism. 
Qualitative results on the predicted uncertainty are reported in \Cref{sec:variance}. As discussed in the paper, the proximity channel is zero over large portions of the BEV grid, since meaningful proximity values are only defined around road-supported regions.
As a result, directly averaging the Laplacian \(\beta\)-NLL over all BEV locations can bias the optimization towards background pixels: the model may reduce the loss by predicting near-zero proximity with very small scale on the background, while leaving road regions weakly supervised and associated with high uncertainty.
To further mitigate this issue beyond the \(\beta\)-weighting term, we introduce a proximity weighting scheme that emphasizes locations with high ground truth proximity values.
The resulting loss is defined as:
\begin{equation}
\mathcal{L}_{L}
=
\frac{1}{|\Omega|}
\sum_{p \in \Omega}
\left(
\log(2 b_p)
+
\frac{\bigl|\hat{\mu}_p-\mu_p\bigr|}{b_p}
\right)
\cdot
\lfloor b_p\rfloor^{\beta}
\cdot
w^{\mathrm{prox}}_{p},
\end{equation}
where \(p \in \Omega\) indexes valid BEV locations, \(\hat{\mu}_p\) is the predicted proximity mean, \(\mu_p\) is the ground truth proximity, and \(b_p\) is the Laplacian scale obtained from the predicted log-scale.
The operator \(\lfloor \cdot \rfloor\) stops gradients through the \(\beta\)-weighting factor, as in the original \(\beta\)-NLL formulation \cite{seitzer2022pitfalls}.
The proximity weight is computed as:
\begin{equation}
w^{\mathrm{prox}}_{p}
=
w_{\min}
+
(1-w_{\min})\,\sigma\!\bigl(\alpha\,(\mu_p-\tau)\bigr).
\end{equation}
where \(\sigma(\cdot)\) denotes the sigmoid function, \(w_{\min}\) is the minimum weight assigned to background regions, \(\tau\) controls the proximity threshold around which the transition occurs, and \(\alpha\) controls the sharpness of the transition.

\input{supplementary/tables/losses_params}

%% file: supplementary/tables/losses_params.tex
\begin{table}[t]
\centering
\small
\label{tab:road_field_loss_hyperparams}

\begin{tabular*}{\columnwidth}{@{\extracolsep{\fill}} l c l}
\toprule
Parameter & Value & Description \\
\midrule
\(w_R\) & 5.0 & \(\mathcal{L}_R\) loss weight \\
\(w_G\) & 2.0 & \(\mathcal{L}_G\) loss weight \\
\(w_L\) & 1.0 & \(\mathcal{L}_L\) loss weight \\
\(\beta\) & 0.5 from \cite{seitzer2022pitfalls, gu2024producing} & \(\beta\)-NLL \(\beta\) param. \\
\(w_{\min}\) & 0.1 & Background weight \\
\(\tau\) & 0.2 & Prox. gate threshold \\
\(\alpha\) & 20.0 & Prox. gate sharpness \\
\bottomrule
\end{tabular*}
\caption{
Hyperparameters used for lane field supervision. Values are selected empirically based on validation performance.
}
\end{table}

%% file: supplementary/sec/4_variance.tex
\section{Qualitative Results on Field Variance}
\label{sec:variance}
To better understand the role of the predicted log-scale in proximity estimation, we report qualitative examples of the resulting uncertainty map.
High variance regions indicate locations where the model is uncertain about the predicted lane structure.
This uncertainty provides a reliability cue for downstream modules, allowing feature modulation to account for uncertain predictions and deformable attention guidance to avoid unreliable regions.
The examples in Figures \ref{fig:78_variance}, \ref{fig:1141_variance} and \ref{fig:90_variance} visualize the variance of the proximity prediction, derived from the predicted log-scale.
Given the predicted log-scale \(s_p=\log (b_p)\), the Laplacian scale is obtained as
\(
b_p = \exp(s_p),
\)
and the corresponding variance is
\(
\operatorname{Var}(p) = 2b_p^2 = 2\exp(2s_p).
\)

%% file: supplementary/sec/5_attention.tex
\section{Analysis of Guided Deformable Attention}
\input{supplementary/tables/attention_gates}
\input{supplementary/tables/attention_metrics}
In this section, we provide additional details on the deformable attention guidance mechanism. As described in the main paper, the deformable attention guidance is controlled by three gates: an anisotropy gate, an uncertainty gate, and a learnable per-head gate.
Each gate produces a scalar value in \([0,1]\) through a sigmoid function.

The anisotropy gate activates the guidance only where the local lane direction is reliable:
\begin{equation}
g_C(p)
=
\sigma\left(\alpha_C\left(C(p)-\tau_C\right)\right),
\end{equation}
where \(\mathbf{C}(p)\) is the direction confidence channel. The uncertainty gate accounts for both predicted lane proximity and its associated uncertainty.
The head predicts the proximity mean \(\hat{\mu}(p)\) and the Laplacian log-scale \(s_p\). 
From \(s_p\), we compute the Laplacian scale \(b_p=\exp(s_p)\) and the corresponding standard deviation as
\[
\operatorname{std}(p)
=
\sqrt{\operatorname{Var}(p)}
=
\sqrt{2}\,b_p
=
\sqrt{2}\exp(s_p).
\]
We then define the uncertainty gate as
\begin{equation}
g_U(p)
=
\sigma\left(
\alpha_U
\left(
(\hat{\mu}(p)-\operatorname{std}(p))-\tau_U
\right)
\right).
\end{equation}
This gate is high only when the model predicts strong lane proximity with low uncertainty. Finally, the learnable per-head gate allows each attention head to control how strongly it uses the LF guidance. 
For head \(h\), we define
\begin{equation}
g_H(h)
=
\sigma\left(\alpha_H \cdot \theta_h\right),
\end{equation}
where \(\theta_h\) is a learnable scalar parameter and \(\alpha_H\) controls the steepness of the sigmoid. \Cref{tab:attention_gates} reports the hyperparameter values used for the gating functions. 
In Figures~\ref{fig:deform}, \ref{fig:deform_intersection} and \ref{fig:high_variance_deform}, we show qualitative examples of how the proposed mechanism modifies the deformable attention sampling pattern in LaneSegNet~\cite{li2023lanesegnet}. Note that these examples are selected from an attention head whose learnable head gate converges close to one. 
Thus, when the direction confidence and uncertainty gates are also active, the mechanism applies the maximum stretch. \Cref{tab:attention_metrics} provides a quantitative comparison between the original Lane Attention module of LaneSegNet and our guided variant. We evaluate three properties of the sampling locations: the alignment error between their principal axis and the ground-truth lane direction, the anisotropy ratio, and the spatial coverage per reference point. \Cref{fig:angle_error_all_heads_mean,fig:anisotropy_all_heads_mean,fig:area_all_heads_mean} report the corresponding metric distributions, computed on 1k validation samples.

%% file: supplementary/tables/attention_gates.tex

\begin{table}[t]
\centering
\small
\setlength{\tabcolsep}{4pt}
\renewcommand{\arraystretch}{1.05}

\newcolumntype{L}{>{\raggedright\arraybackslash}X}
\newcolumntype{R}{>{\centering\arraybackslash}X}

\begin{tabularx}{\columnwidth}{L R}
\toprule
Gate & Parameters \\
\midrule
Anisotropy gate 
& \(\alpha_C=10.0,\ \tau_C=0.3\) \\

Uncertainty gate 
& \(\alpha_U=10.0,\ \tau_U=0.3\) \\

Learnable per-head gate 
& \(\alpha_H=2.0\) \\

Max Prior Stretch 
& \(\Delta_{\parallel}=4.0\) \\
\bottomrule
\end{tabularx}

\caption{
Parameters of the gates used in the deformable attention guidance mechanism.
}
\label{tab:attention_gates}
\end{table}

%% file: supplementary/tables/attention_metrics.tex
\begin{table}[t]
\centering
\small
\setlength{\tabcolsep}{5pt}
\renewcommand{\arraystretch}{1.12}

\begin{tabularx}{\columnwidth}{Xccc}
\hline
\textbf{Method} 
& \shortstack{$\mathbf{E}_{align}\downarrow$}
& \shortstack{\textbf{Anisotropy}}
& \shortstack{\textbf{Area}} \\
\hline
Lane Attention~\cite{li2023lanesegnet}
& 35.3° & 13\% & 33.2$\mathrm{m}^2$ \\

Guided Attention (\textbf{Ours})
& 11.4° & 62\% & 78.2$\mathrm{m}^2$ \\
\hline
\end{tabularx}

\caption{
Quantitative comparison between the original Lane Attention of LaneSegNet~\cite{li2023lanesegnet} and our conditioned variant.
}
\label{tab:attention_metrics}
\end{table}

%% file: supplementary/sec/6_modulation.tex
\section{Analysis of BEV Features Modulation}
We analyze how the proposed modulation block affects the BEV feature representation. 
\Cref{fig:mod_01} reports three complementary analyses before and after applying LF modulation. 
First, we visualize the BEV features through PCA, observing that the modulated representation exhibits a clearer lane structure. This suggests that the modulation block injects geometric information from the LF into the latent BEV space. We further quantify this effect by clustering BEV locations in feature space using K-means and measuring the average ground truth proximity channel within each cluster. 
Before modulation, the clusters show similar average proximity values, indicating that the original BEV features are only weakly organized according to lane support. 
After modulation, the clusters become more clearly separated with respect to proximity, and the background cluster collapses toward zero, showing that road-supported and non-road regions are better distinguished in the refined representation.
Finally, we approximate the BEV feature manifold using UMAP~\cite{mcinnes2018umap}, reducing each BEV feature vector from $C$ dimensions to two dimensions. 
By selecting BEV locations with road evidence through a proximity threshold, we observe that road-supported locations become more compact in the embedded space after modulation. 
Overall, these analyses indicate that LF modulation enriches BEV features with structured geometric cues and makes the representation more sensitive to the underlying road layout.

%% file: supplementary/sec/7_qualitative.tex
\section{Additional Qualitative Results}

We provide additional qualitative results on OpenLane-V2 to complement the quantitative evaluation. 
Compared with the LaneSegNet baseline, our method produces more precise and structurally consistent map predictions, particularly in scenes with elongated structures. 
As shown in Figures \ref{fig:qual_01} and \ref{fig:qual_02} LF conditioning helps preserve long-range geometric continuity and reduces fragmented or locally inconsistent predictions. \Cref{fig:fail_01} shows a failure case involving a rare roundabout scenario, \Cref{fig:fail_02} shows a failure case in which the model misses a pedestrian crossing. 

%% file: supplementary/sec/8_inference.tex
\section{Inference Speed and Model Size}
\input{supplementary/tables/inference}
\input{supplementary/tables/sizes}
We report the inference speed of the baselines and their Lane Field enhanced variants in FPS and TFLOPs. 
All measurements are obtained on an NVIDIA RTX 4000 Ada Generation GPU and should be interpreted as reference throughput values, since absolute speed can depend on hardware and memory-bandwidth constraints. 
As shown in \Cref{tab:inference_speed}, our framework introduces only a minor FPS / TFLOPs overhead compared with the corresponding baselines.
\Cref{tab:sizes} reports the parameter count of each baseline and its LF-enhanced variant.

%% file: supplementary/tables/inference.tex
\begin{table}[t]
\centering
\small
\setlength{\tabcolsep}{2.5pt}
\renewcommand{\arraystretch}{1.12}

\begin{tabularx}{\columnwidth}{@{}Xccccc@{}}
\hline
\textbf{Baseline}
& \textbf{Model}
& \textbf{+SPADE}
& \textbf{+Attn.}
& \textbf{+All} \\
\hline

LaneSegNet~\cite{li2023lanesegnet}
& 14.6 / 1.3
& 13.9 / 1.5
& 13.9 / 1.5
& 13.9 / 1.5 \\

MapTRv2~\cite{liao2025maptrv2}
& 33.2 / 1.2
& 29.9 / 1.4
& 29.9 / 1.4
& 29.9 / 1.4 \\

TopoPoint~\cite{fu2026topopoint}
& 13.8 / 1.4
& 13.0 / 1.6
& 13.0 / 1.6
& 13.0 / 1.6 \\

\hline
\end{tabularx}

\caption{
Breakdown of the computational overhead introduced by the proposed Lane Field modules across different baselines. Results are reported as FPS / TFLOPs, respectively.
}
\label{tab:inference_speed}
\end{table}

%% file: supplementary/tables/sizes.tex
\begin{table}[t]
\centering
\small
\setlength{\tabcolsep}{3pt}
\renewcommand{\arraystretch}{1.12}

\begin{tabularx}{\columnwidth}{Xccccc}
\hline
\textbf{Baseline}
& \textbf{Model}
& \textbf{+Head}
& \textbf{+SPADE}
& \textbf{+Attn.}
& \textbf{+All} \\
\hline

LaneSegNet~\cite{li2023lanesegnet}
& 45.4M
& 4.8M
& 2.4M
& $\approx$0
& 52.6M \\

MapTRv2~\cite{liao2025maptrv2}
& 40.6M
& 4.8M
& 2.4M
& $\approx$0
& 47.8M \\

TopoPoint~\cite{fu2026topopoint}
& 53.6M
& 4.8M
& 2.4M
& $\approx$0
& 60.8M \\
\hline
\end{tabularx}

\caption{
Breakdown of the parameter overhead introduced by the proposed Lane Field modules across different baselines.
}
\label{tab:sizes}
\end{table}

%% file: supplementary/sec/9_gradients.tex
\section{Ablation on Gradient Flow}
\input{supplementary/tables/gradients}

\Cref{tab:gradients} reports the results for each gradient-stopping configuration. AugMapNet \cite{monninger2026augmapnet} uses Stop-grad. In \& Out.

%% file: supplementary/tables/gradients.tex
\definecolor{HeaderGray}{gray}{1.00}
\definecolor{ResultGray}{gray}{0.86}

\begin{table}[t]
\centering
\small
\setlength{\tabcolsep}{4pt}
\renewcommand{\arraystretch}{1.15}

\begin{tabularx}{\columnwidth}{
l
>{\centering\arraybackslash}X
>{\centering\arraybackslash}X
>{\centering\arraybackslash}X
>{\centering\arraybackslash}X
}
\hline
\textbf{Gradients conf.} & \textbf{mAP}\;$\uparrow$ & \textbf{AP$_{ls}$}\;$\uparrow$ & \textbf{AP$_{ped}$}\;$\uparrow$ & \textbf{TOP$_{lsls}$}\;$\uparrow$ \\
\hline

Baseline
& 32.1
& 32.2
& 31.9
& 25.5 \\

Stop-grad. In \& Out
& 34.1
& 34.2
& 34.0
& 26.7 \\

Stop-grad. In
& 34.8
& 34.4
& 35.2
& 27.0 \\

Stop-grad. Out
& 33.3
& 32.3
& 34.3
& 27.2 \\

\rowcolor{ResultGray}
\textbf{Fully attached}
& \textbf{36.1}
& \textbf{35.8}
& \textbf{36.4}
& \textbf{28.4} \\

\hline
\end{tabularx}

\caption{Gradient-flow configurations of our framework. Stop-grad. In detaches the BEV features before the field head, while Stop-grad. Out detaches the LF at the head output.}
\label{tab:gradients}
\end{table}

%% file: supplementary/sec/big_images.tex







\begin{figure*}[!t]
    \centering

    \includegraphics[width=0.32\textwidth]{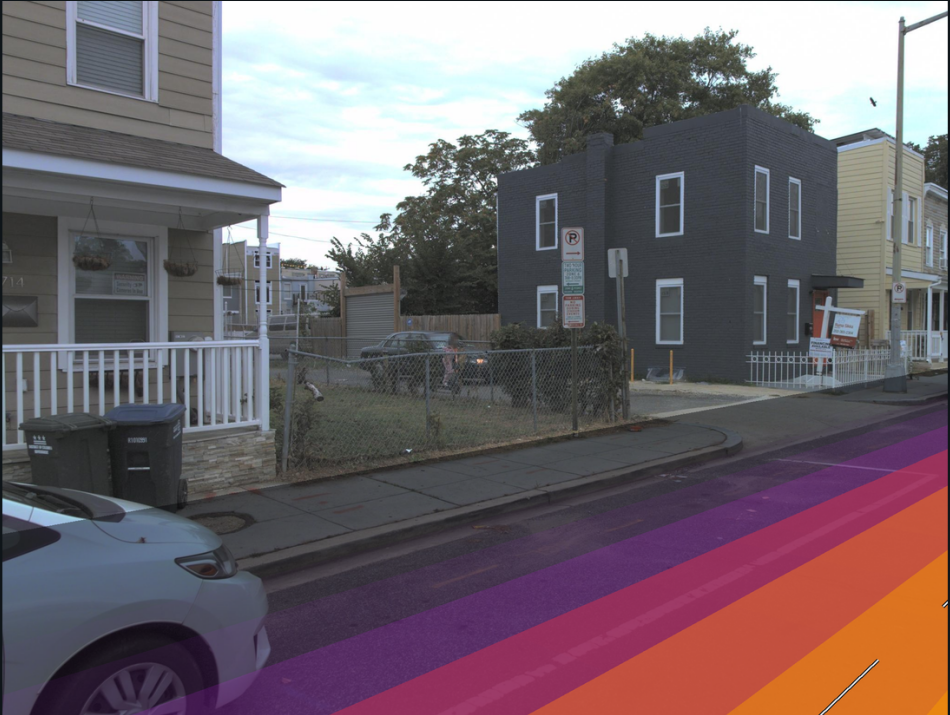}
    \hfill
    \includegraphics[width=0.32\textwidth]{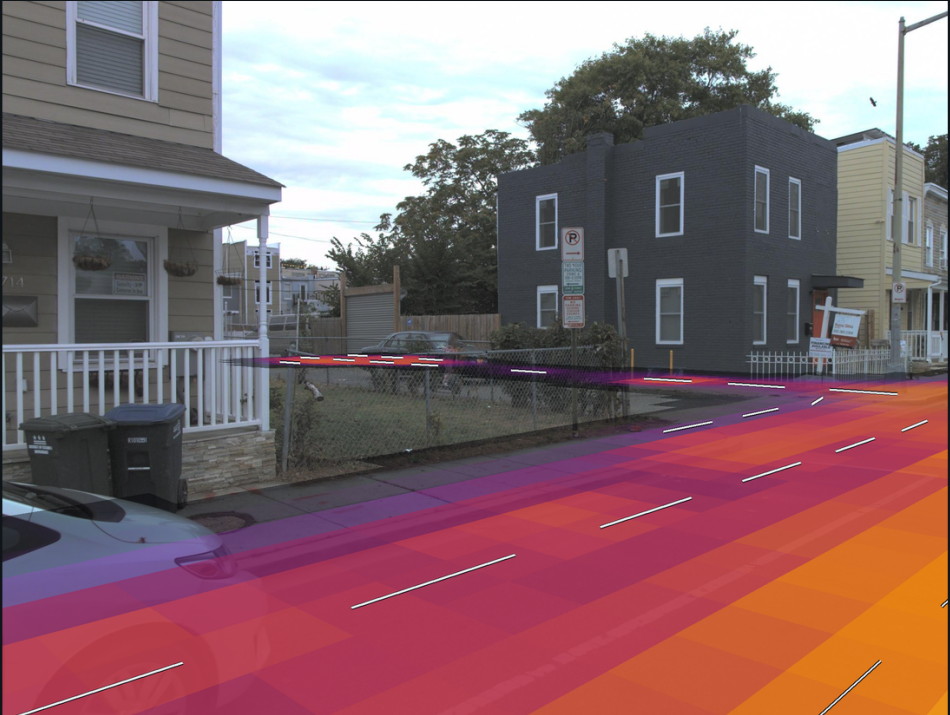}
    \hfill
    \includegraphics[width=0.32\textwidth]{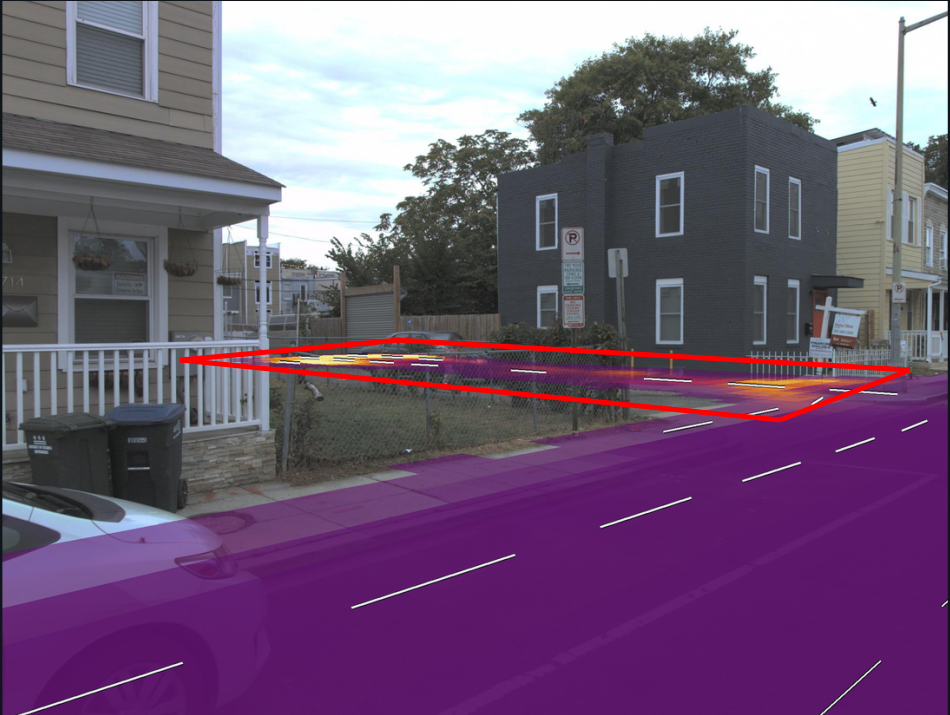}

\caption{
Predicted variance as an uncertainty cue for lane field guidance.
The driveway is not annotated in the ground truth (left).
Although the model predicts high proximity in that region (center), it also predicts high variance (right), indicating uncertainty and allowing downstream modules to reduce the use of the lane field there.
}
    \label{fig:78_variance}
\end{figure*}
\begin{figure*}[!t]
    \centering

    \includegraphics[width=0.32\textwidth]{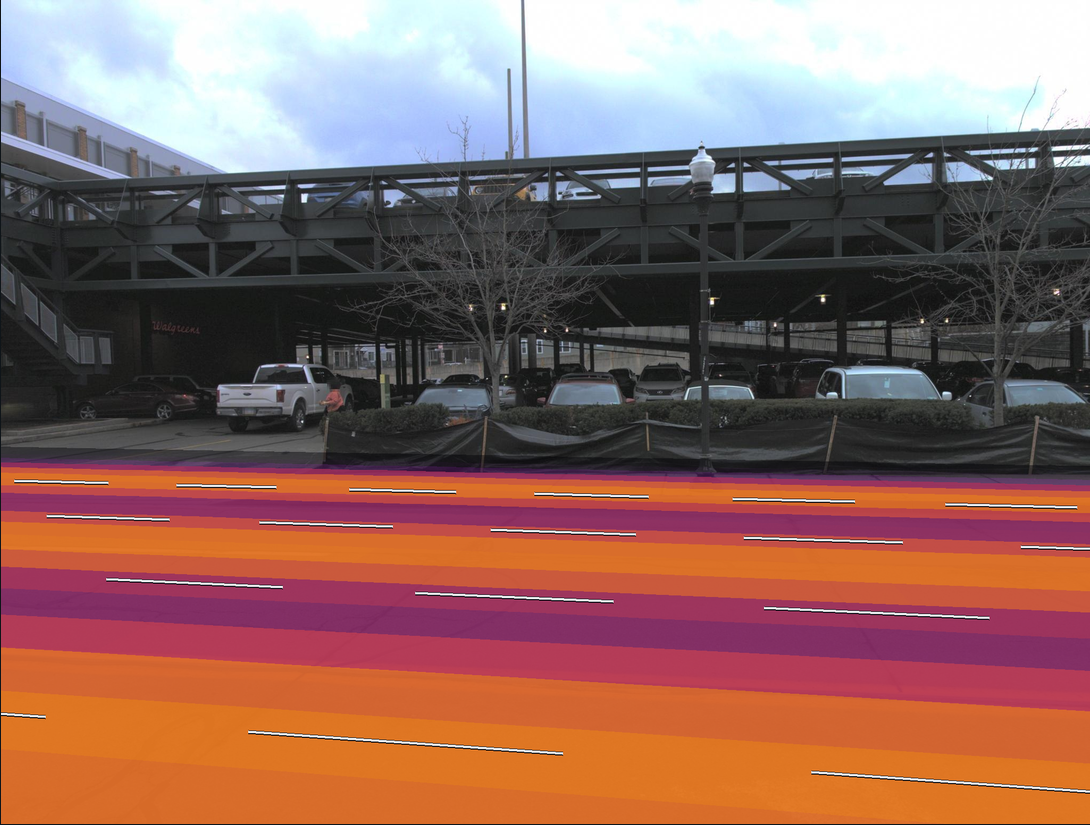}
    \hfill
    \includegraphics[width=0.32\textwidth]{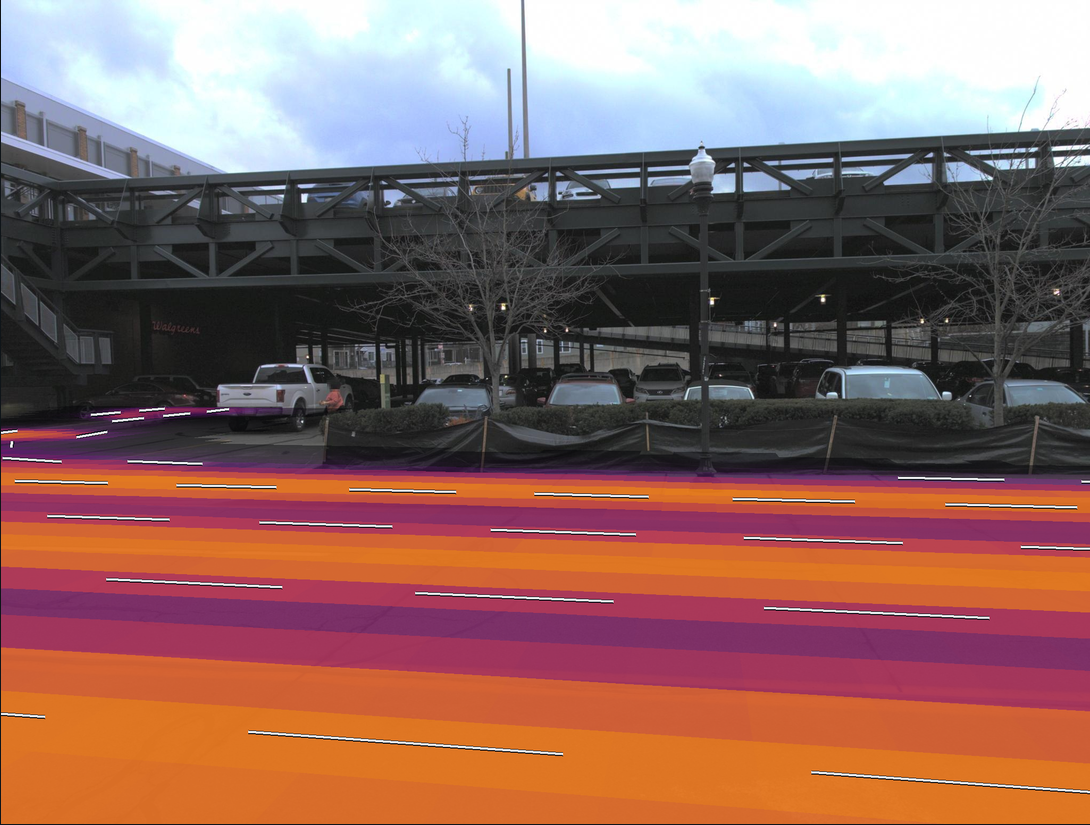}
    \hfill
    \includegraphics[width=0.32\textwidth]{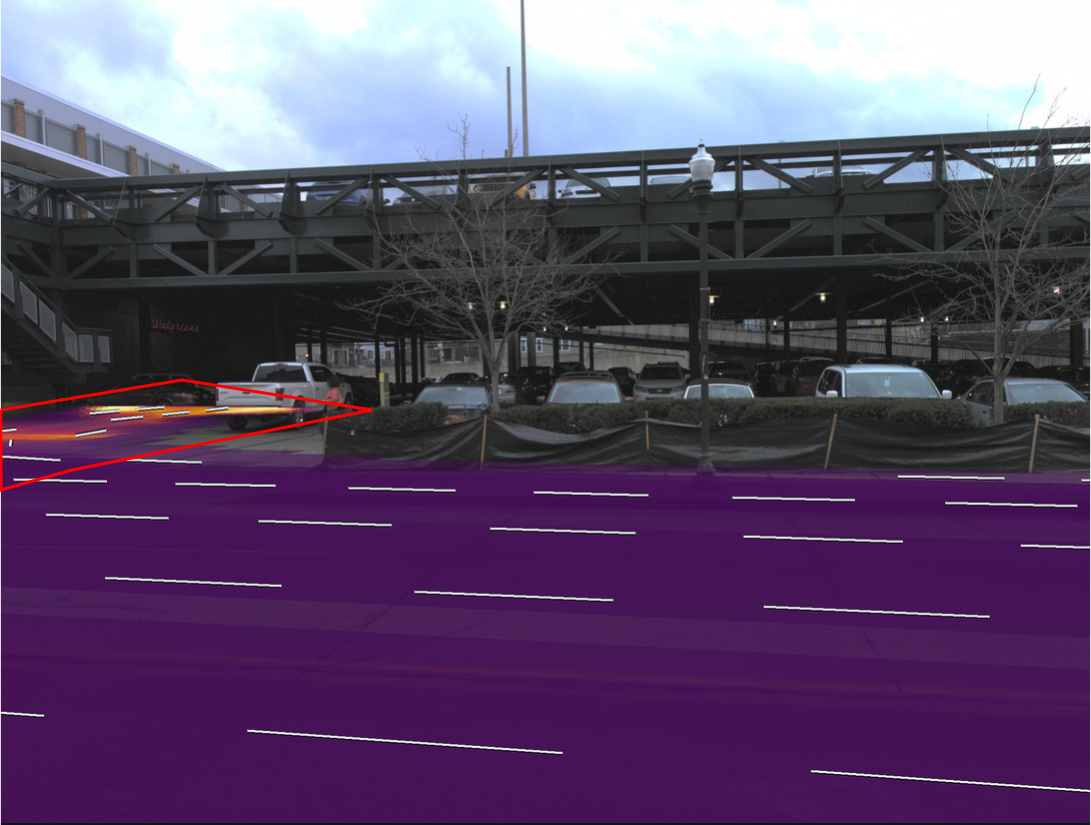}

\caption{
The parking lot entrance is not annotated in the ground truth (left).
Although the model predicts high proximity in that region (center), it also predicts high variance (right), indicating uncertainty and allowing downstream modules to reduce the use of the lane field there.
}
    \label{fig:1141_variance}
\end{figure*}
  
\begin{figure*}[!t]
    \centering

    \includegraphics[width=0.32\textwidth]{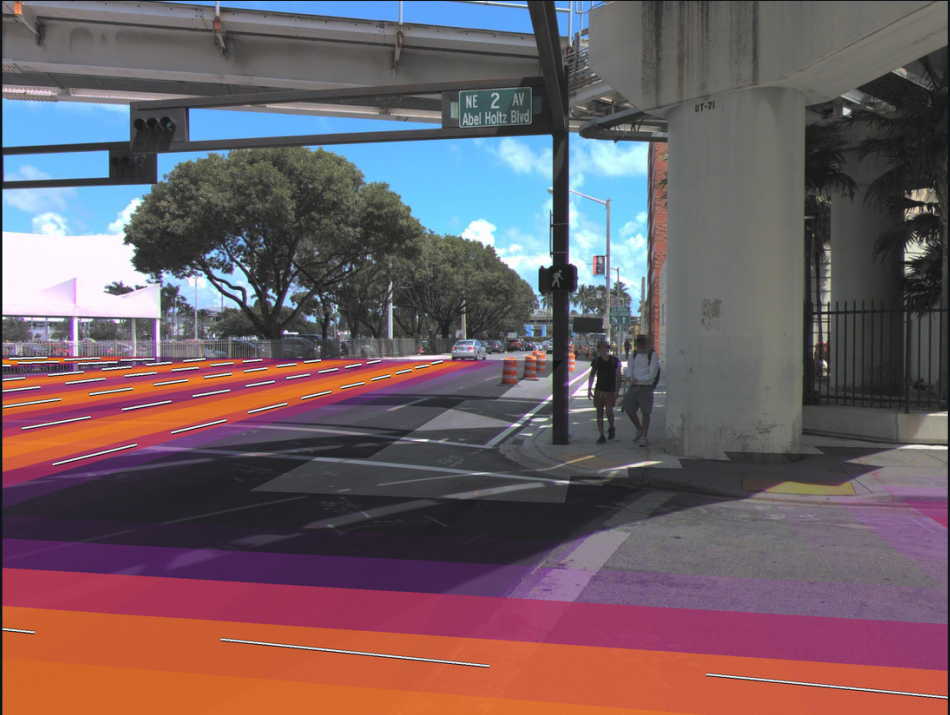}
    \hfill
    \includegraphics[width=0.32\textwidth]{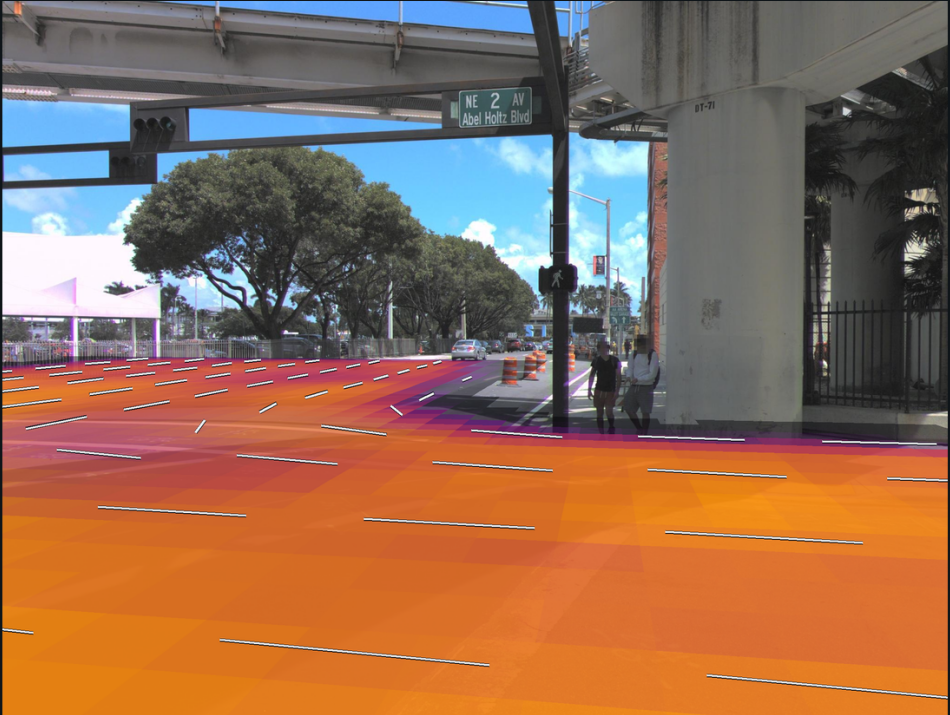}
    \hfill
    \includegraphics[width=0.32\textwidth]{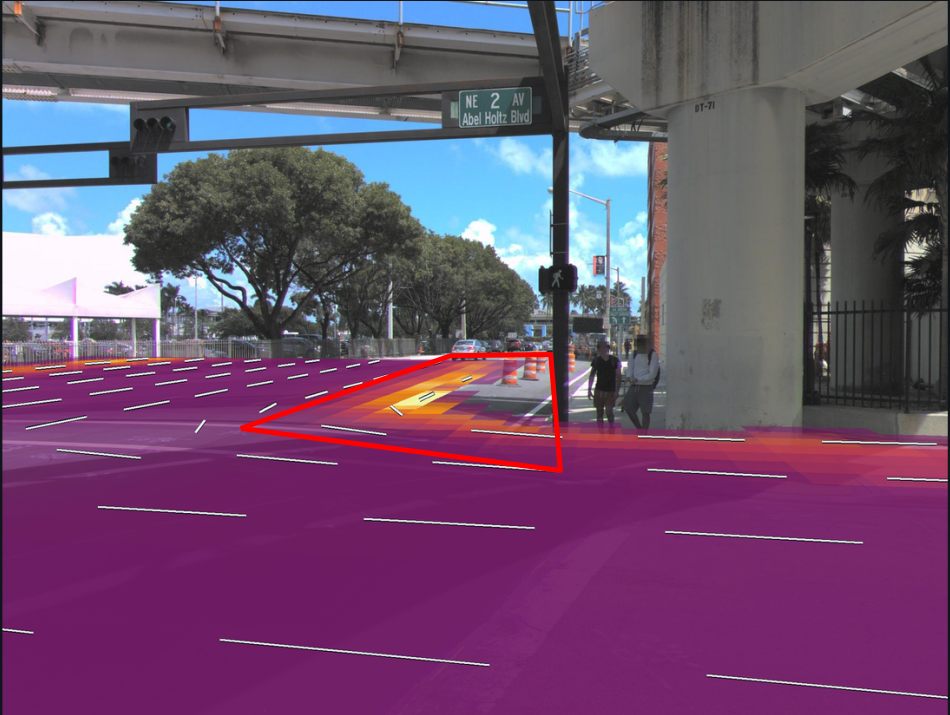}

\caption{
Example of an ambiguous free-space region outside the drivable lanes.
The model assigns high variance to regions where it is uncertain whether the observed area should be treated as drivable area.
}
    \label{fig:90_variance}
\end{figure*}

\clearpage
\begin{figure*}[!t]
    \centering

    \includegraphics[width=0.32\textwidth]{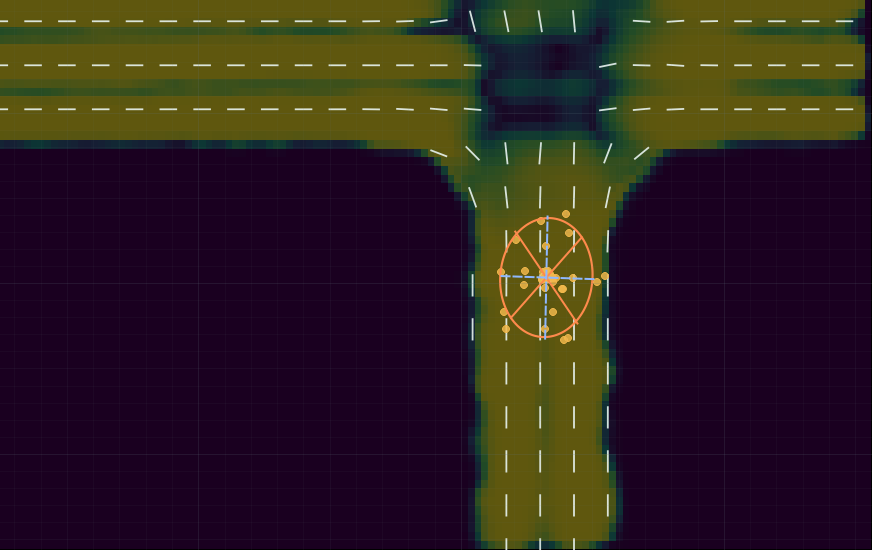}
    \hfill
    \includegraphics[width=0.32\textwidth]{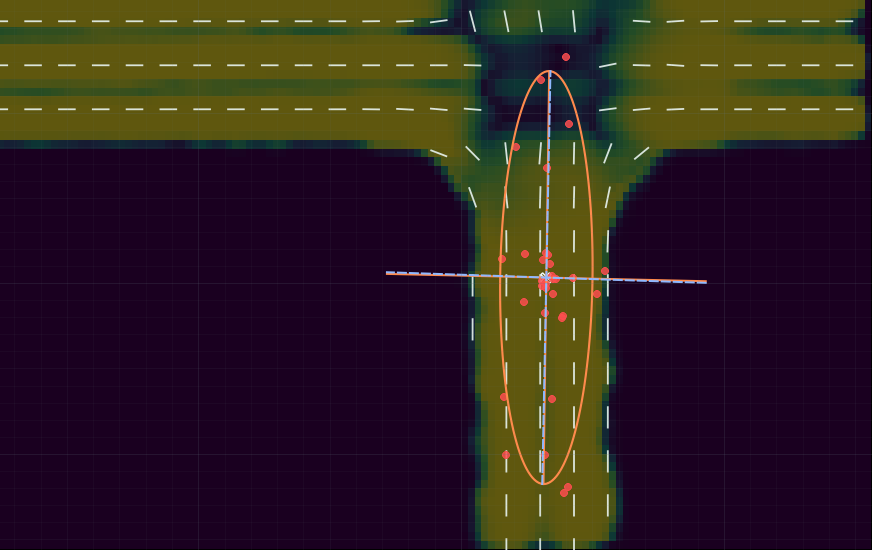}
    \hfill
    \includegraphics[width=0.32\textwidth]{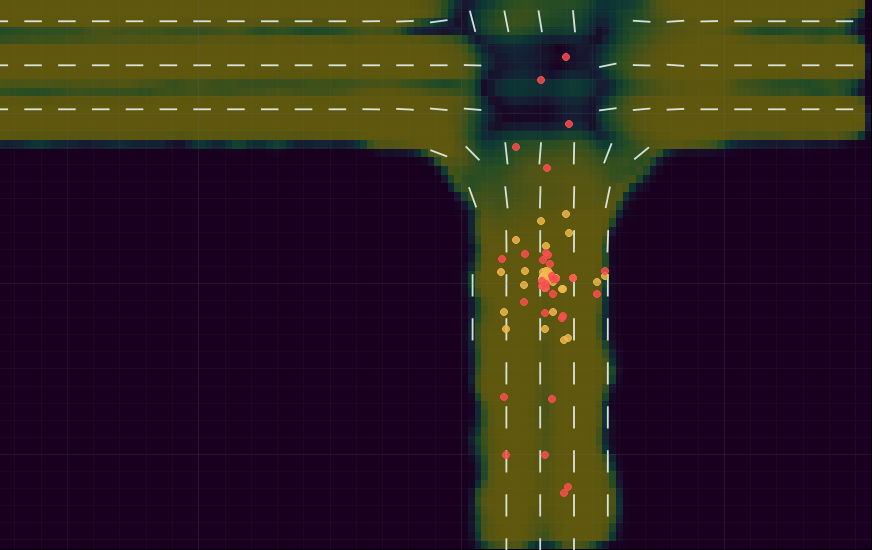}

\caption{
Visualization of the sampling locations associated with a reference point lying on the road at the last decoder layer.
Without guidance (left), the sampling locations follow a mostly local and isotropic pattern, and the first principal component, \ie the direction of maximum variance is misaligned with the local lanes direction.
With lane field guidance (center), the sampling pattern is stretched along the lane structure, and its first principal component becomes well aligned with the local lane direction. The right panel compares the original sampling locations with the guided ones, showing how the offsets are stretched along the local lane direction.}
    \label{fig:deform}
\end{figure*}
\begin{figure*}[!t]
    \centering

    \includegraphics[width=0.32\textwidth]{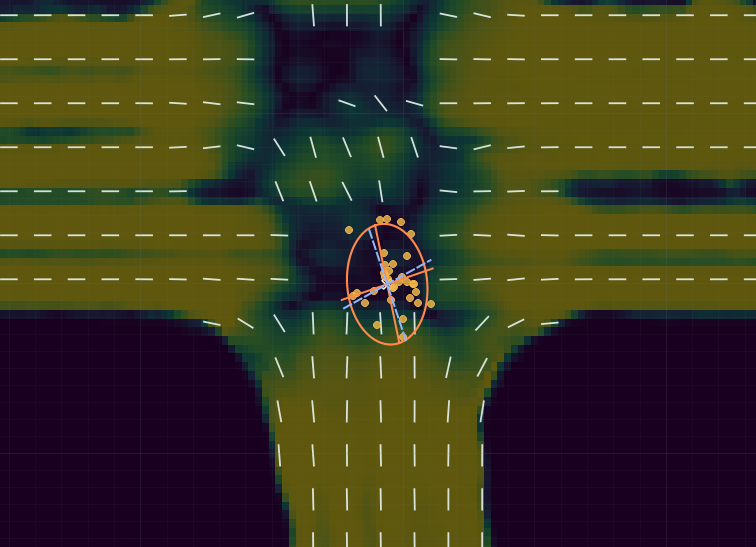}
    \hfill
    \includegraphics[width=0.32\textwidth]{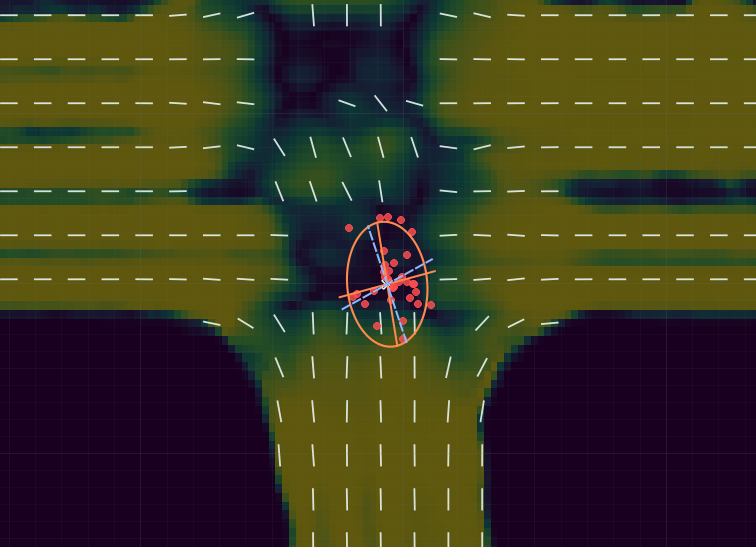}
    \hfill
    \includegraphics[width=0.32\textwidth]{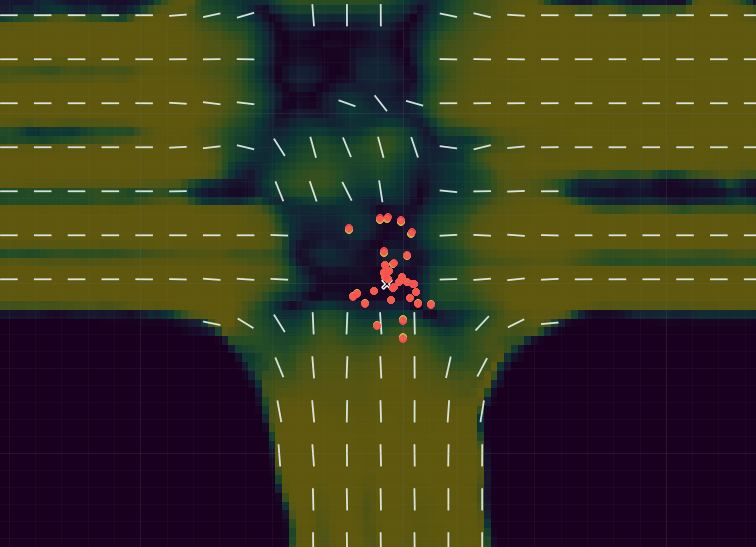}

\caption{
Visualization of the sampling locations associated with a reference point lying at an intersection in the last decoder layer.
Since the direction confidence \(\mathbf{C}(p)\) is close to zero in this ambiguous region, the lane field guidance is suppressed and the sampling pattern remains close to the original learned offsets.
}
    \label{fig:deform_intersection}
\end{figure*}
\begin{figure*}[!t]
    \centering

    \includegraphics[width=0.32\textwidth]{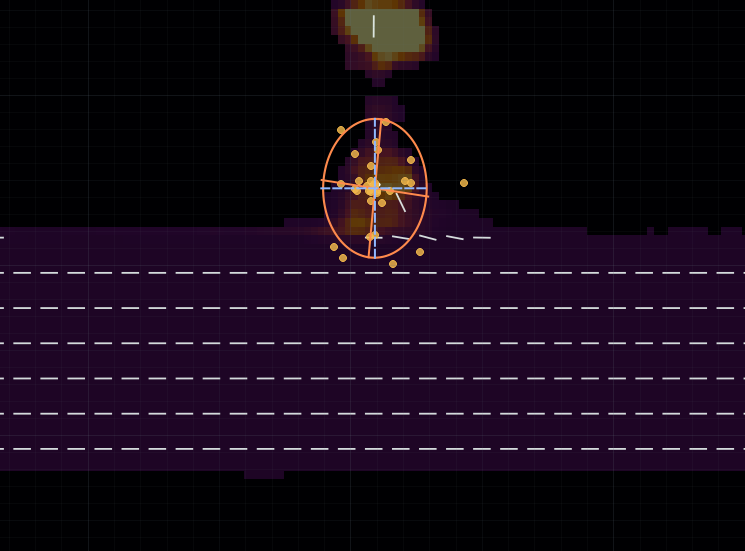}
    \hfill
    \includegraphics[width=0.32\textwidth]{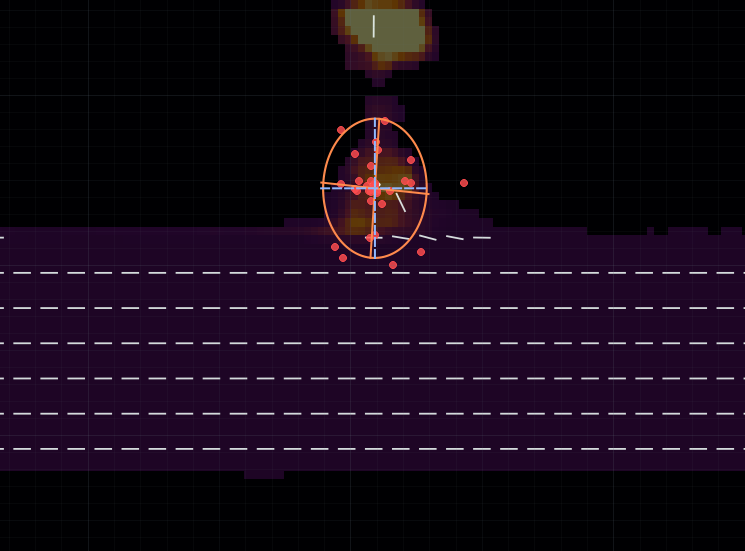}
    \hfill
    \includegraphics[width=0.32\textwidth]{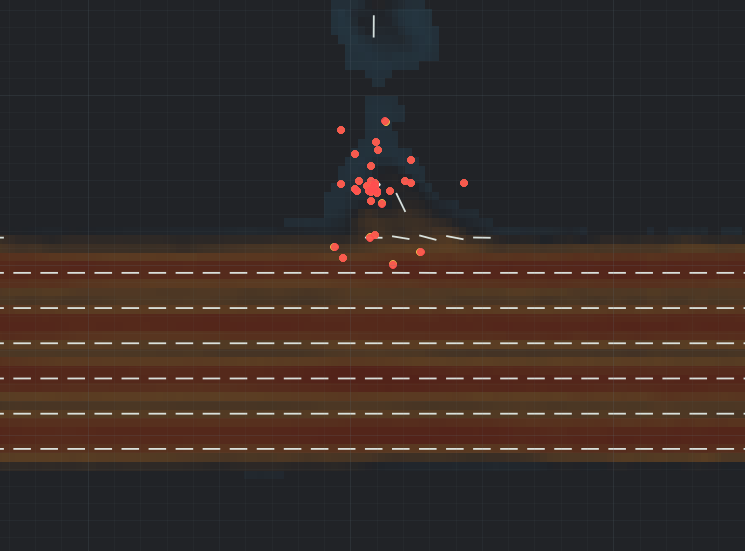}

\caption{
Visualization of the sampling locations associated with a reference point lying in a high variance region, corresponding to the area highlighted in \Cref{fig:1141_variance}, at the last decoder layer.
Since the variance is high in this region, the uncertainty gate suppresses the lane field guidance and the deformable attention sampling remains close to the original learned offsets.
The right panel highlights in red the areas considered reliable by the uncertainty gate.
}
    \label{fig:high_variance_deform}
\end{figure*}

\clearpage
\begin{figure*}[!t]
    \centering

    \includegraphics[width=0.9\textwidth]{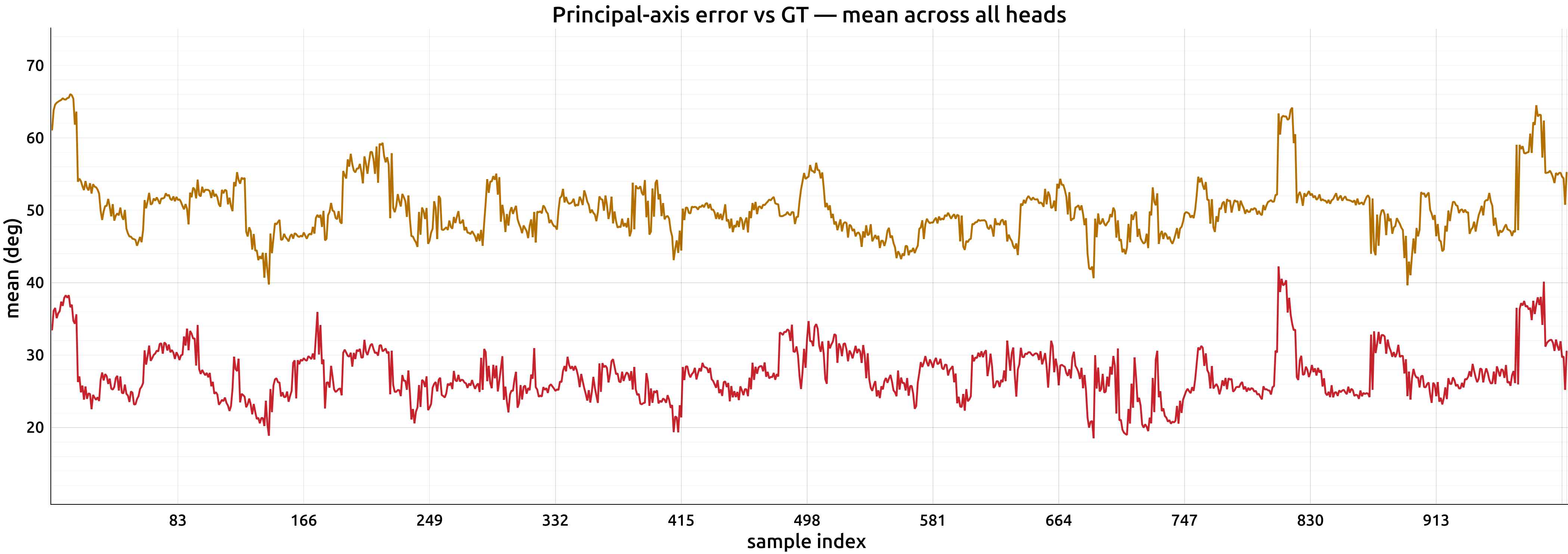}

\caption{
Alignment error (in degrees) between the principal axis of the sampling locations, i.e., the direction of maximum spatial extent, and the ground-truth lane direction. Our conditioning method (red) exhibits substantially better alignment than the baseline sampling locations (orange).
}
    \label{fig:angle_error_all_heads_mean}
\end{figure*}

\begin{figure*}[!t]
    \centering

    \includegraphics[width=0.9\textwidth]{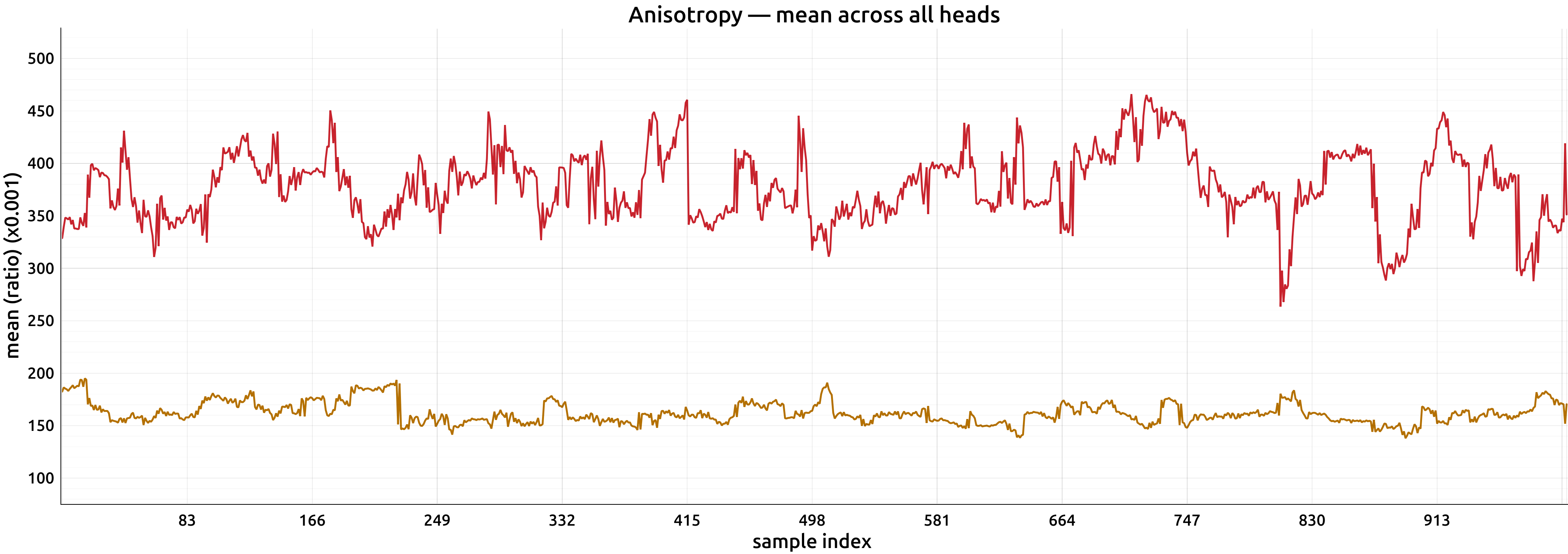}
\caption{
Anisotropy comparison between our conditioned sampling locations (red) and the baseline sampling locations (orange). Our approach exhibits a clearer anisotropic pattern aligned with the lane direction, whereas the baseline sampling locations remain largely isotropic.
}
    \label{fig:anisotropy_all_heads_mean}
\end{figure*}

\begin{figure*}[!t]
    \centering

    \includegraphics[width=0.9\textwidth]{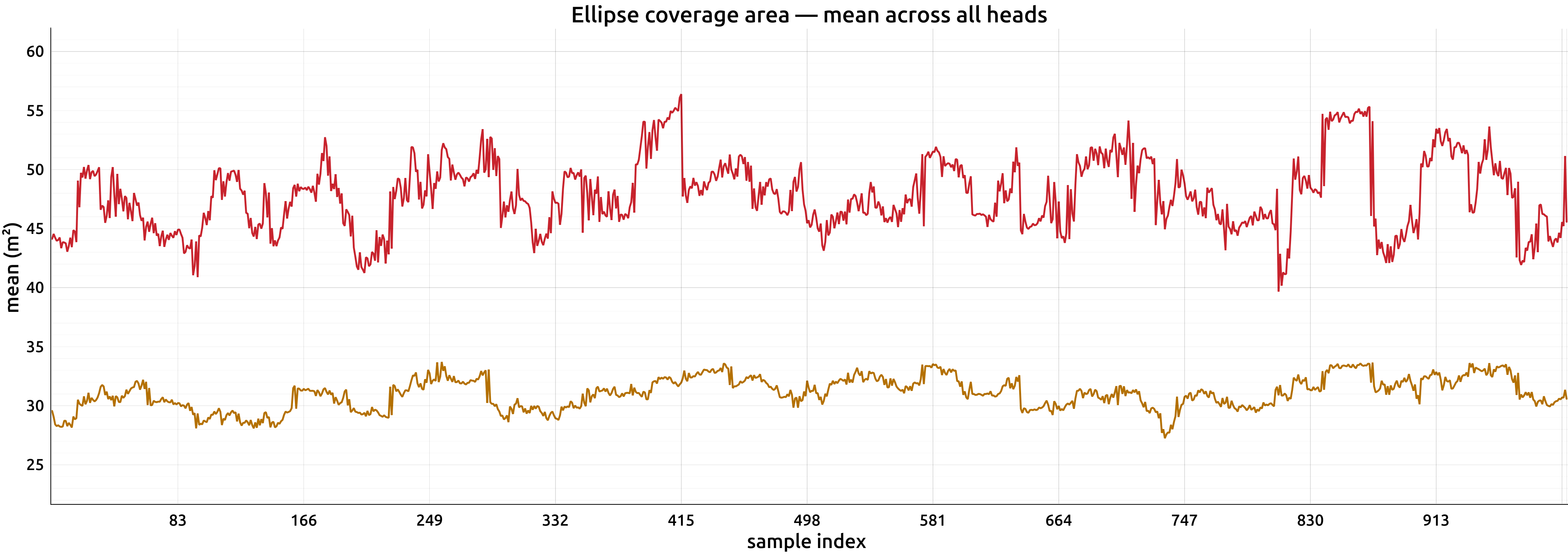}

\caption{
Area comparison (in $\mathrm{m}^2$) between our conditioned sampling locations (red) and the baseline sampling locations (orange). The coverage is computed as the average area covered per reference point. Our method achieves consistently larger spatial coverage than the baseline.
}
    \label{fig:area_all_heads_mean}
\end{figure*}


\clearpage
\begin{figure*}[p]
    \centering
\includegraphics[
        width=\textwidth,
        height=0.93\textheight,
        keepaspectratio
    ]{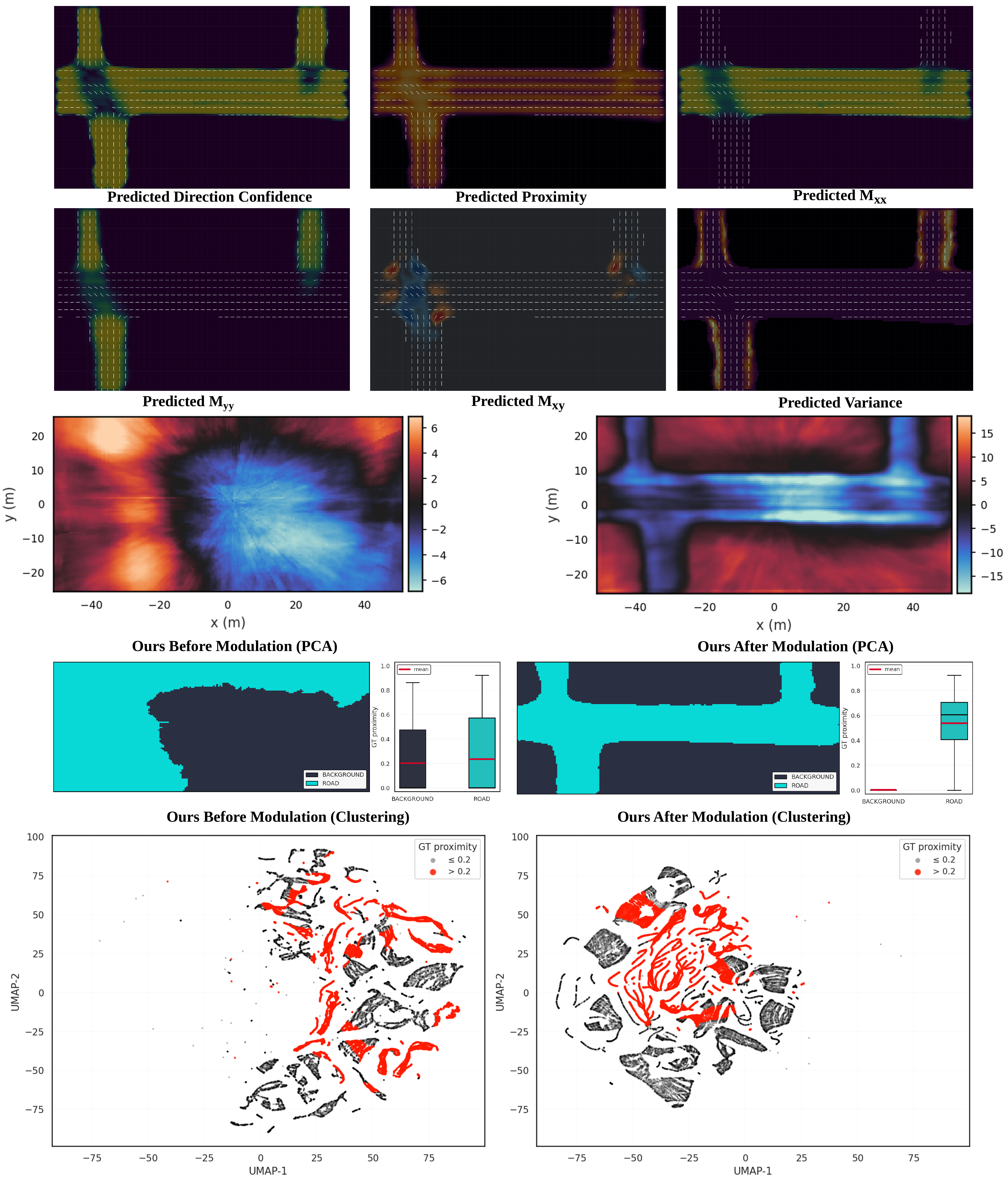}
\caption{
Comparison of BEV features before and after the proposed modulation block. 
The modulated features show a clearer road-structured organization, suggesting that the proposed lane field conditioning effectively enriches the BEV representation with geometric cues.
 }
    \label{fig:mod_01}
\end{figure*}

\clearpage
\begin{figure*}[p]
    \centering
\includegraphics[
        width=\textwidth,
        height=0.93\textheight,
        keepaspectratio
    ]{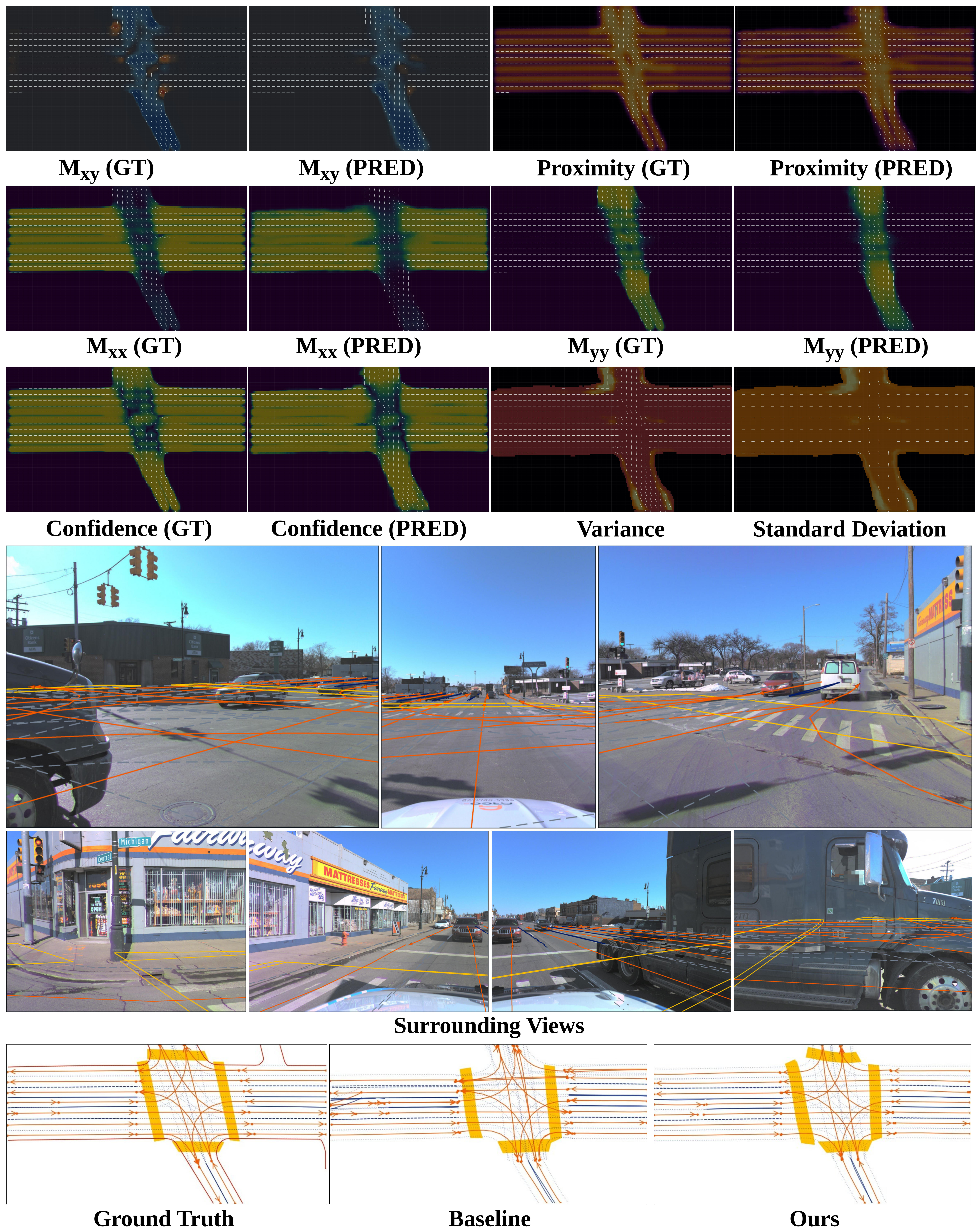}
\caption{
Qualitative comparison on OpenLane-V2.
The LaneSegNet baseline fails to recover a pedestrian crossing and produces inaccurate local predictions, whereas our method correctly detects the crossing and yields a more precise and structurally consistent map.
}
    \label{fig:qual_01}
\end{figure*}

\clearpage
\begin{figure*}[p]
    \centering
\includegraphics[
        width=\textwidth,
        height=0.93\textheight,
        keepaspectratio
    ]{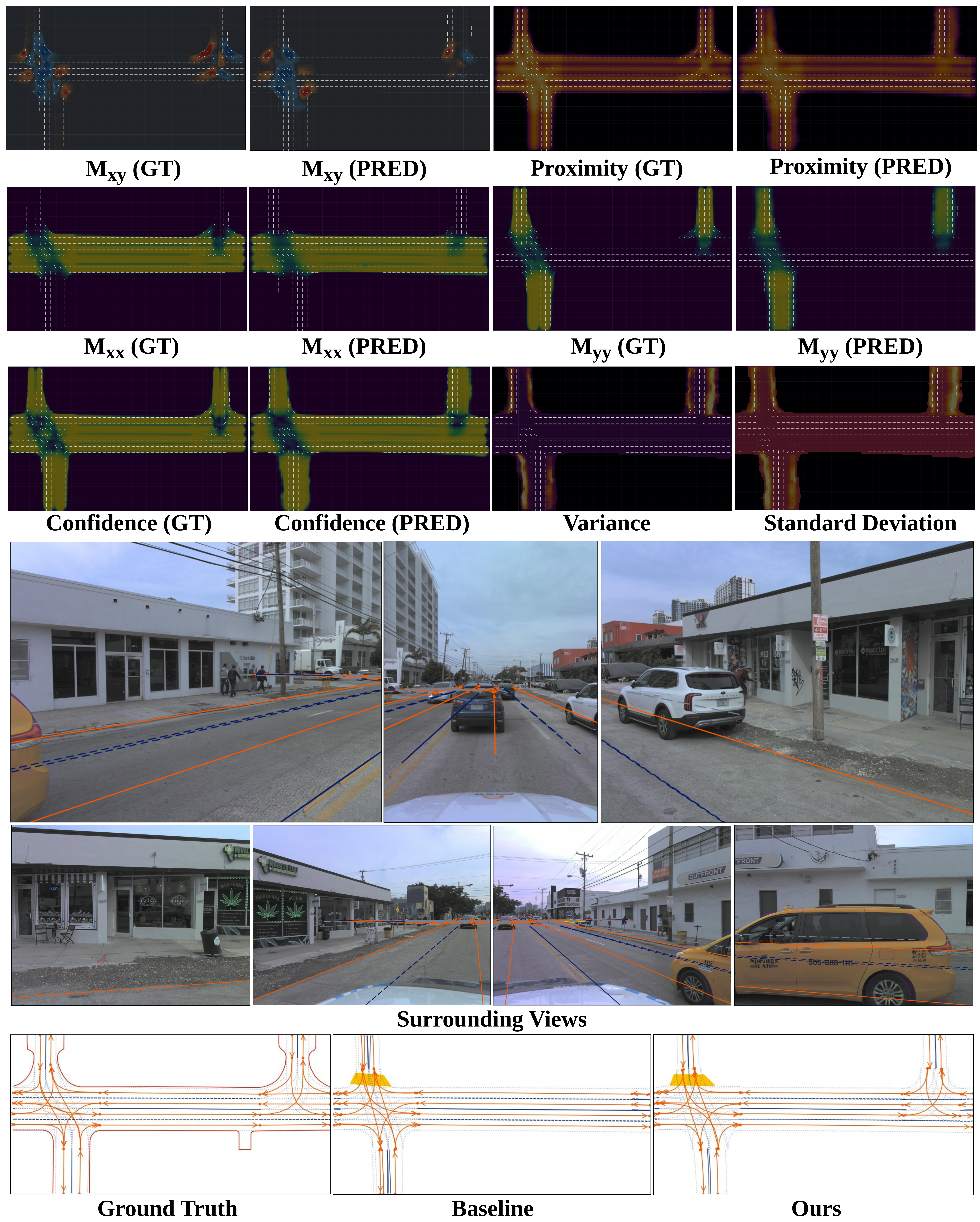}
\caption{
Another comparison on OpenLane-V2 between the LaneSegNet baseline and our method. 
In this example, the baseline fails to recover a distant intersection visible in the HD map, whereas our method preserves its overall structure more accurately. 
We also report the predicted and ground truth lane field channels in BEV to illustrate the geometric cues used by our framework.
}
    \label{fig:qual_02}
\end{figure*}

\clearpage
\begin{figure*}[p]
    \centering
\includegraphics[
        width=\textwidth,
        height=0.93\textheight,
        keepaspectratio
    ]{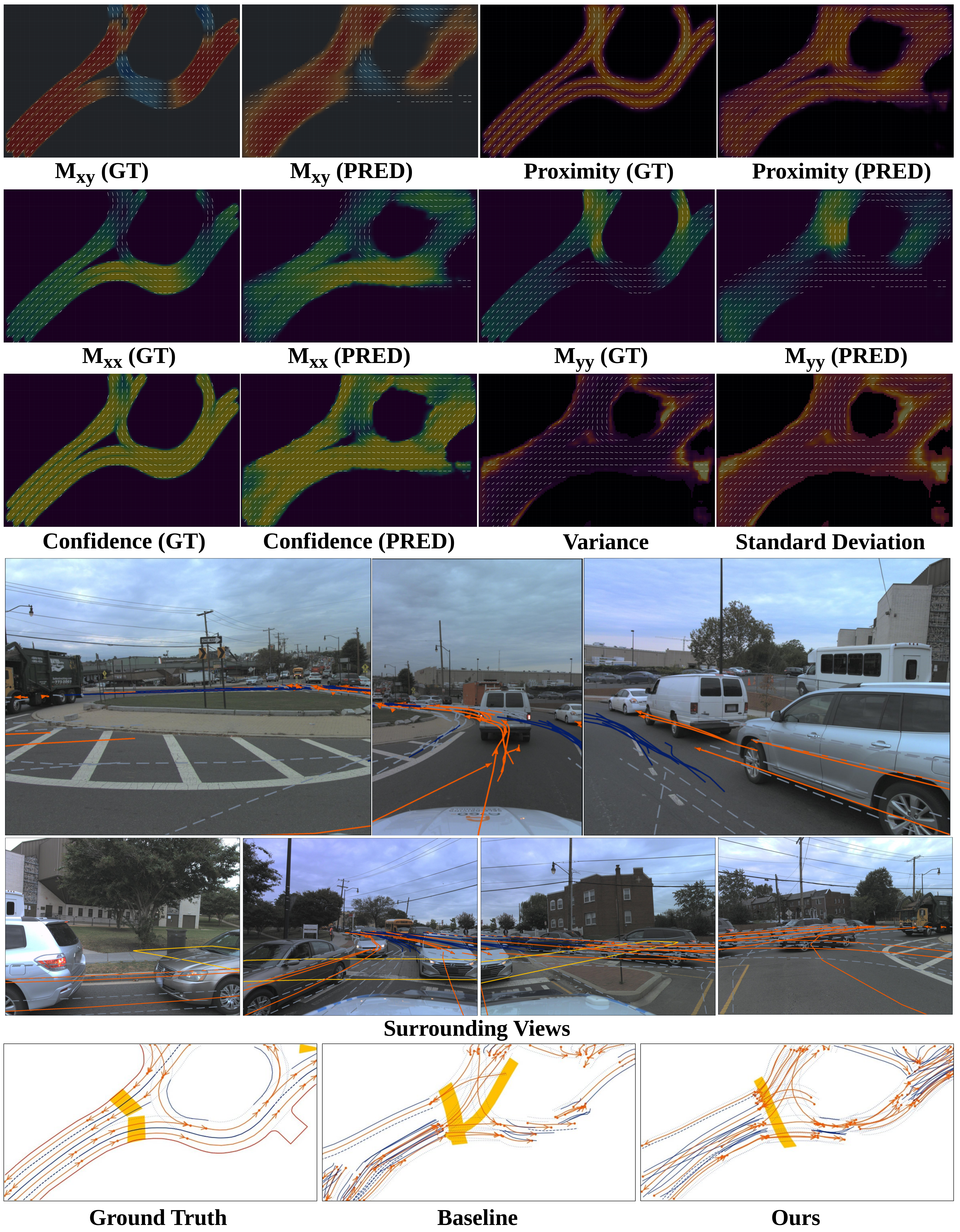}
\caption{
Failure case on OpenLane-V2. 
Due to the rarity of roundabouts in the dataset, both the LaneSegNet baseline and our method struggle to produce an accurate HD map prediction in this scenario. 
Nevertheless, our method shows a slightly better awareness of the overall lane structure.
}
    \label{fig:fail_01}
\end{figure*}

\clearpage
\begin{figure*}[p]
    \centering
\includegraphics[
        width=\textwidth,
        height=0.93\textheight,
        keepaspectratio
    ]{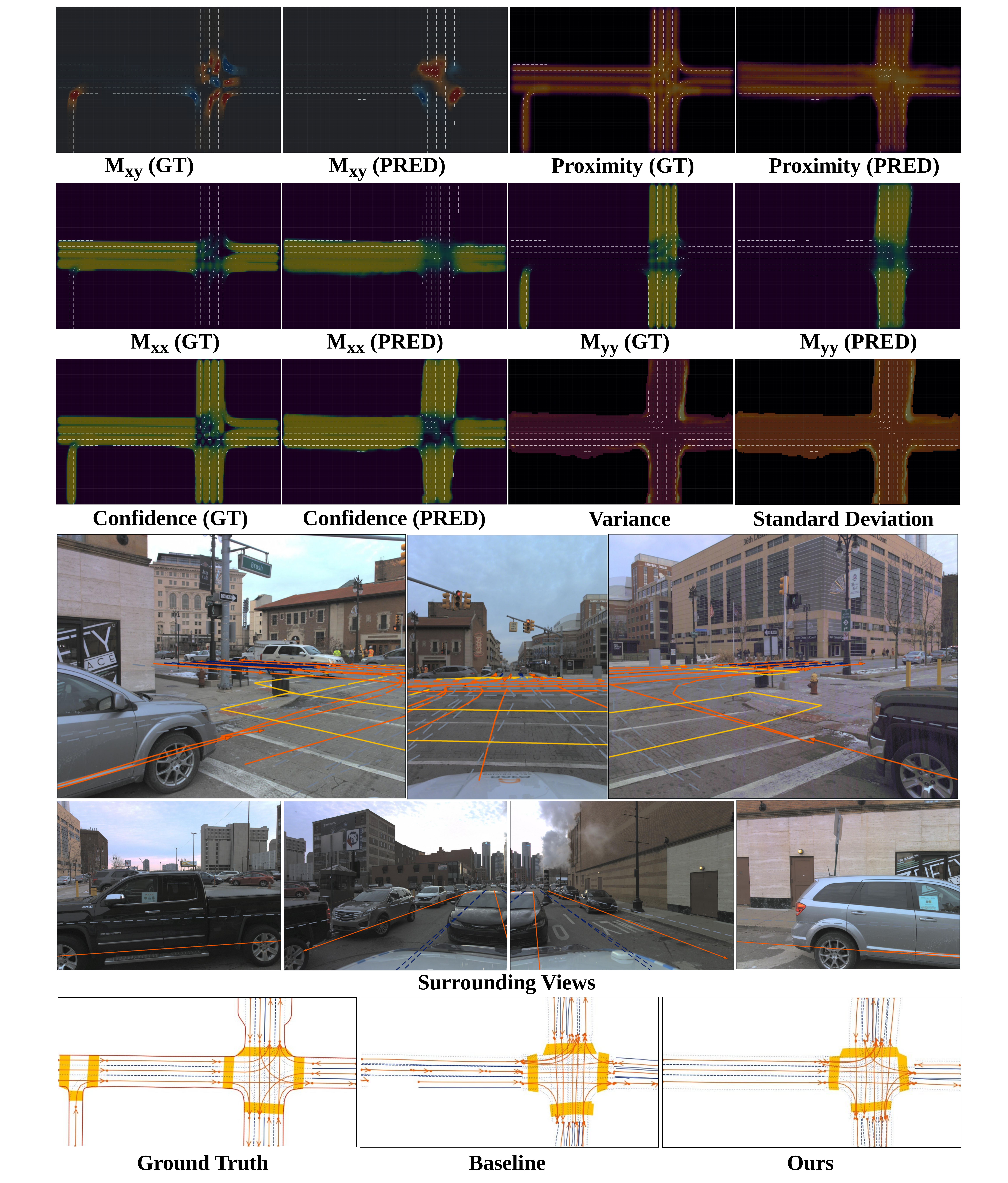}
\caption{
Another failure case on OpenLane-V2. 
Although our method improves the overall lane structure prediction, the lane field does not explicitly encode pedestrian crossing regions, and the model may therefore still miss some crossings in challenging scenes.
}
    \label{fig:fail_02}
\end{figure*}